\documentclass[oneside,10pt]{article}

\usepackage{times,url,mathrsfs,algorithm}
\usepackage{amsmath}
\usepackage{amsfonts}
\usepackage{graphicx}
\usepackage{subfigure}

\begin{document}

\title{ABC-LogitBoost for Multi-class Classification}

\author{ Ping Li \\
       Department of Statistical Science\\
       Faculty of Computing and Information Science\\
       Cornell University\\
       Ithaca, NY 14853\\
       pingli@cornell.edu}
\date{}

\maketitle

\begin{abstract}

We develop {\em abc-logitboost}, based on the prior work on {\em abc-boost}\cite{Proc:ABC_ICML09} and {\em robust logitboost}\cite{Report:Li_Robust-LogitBoost}. Our extensive experiments on a variety of datasets demonstrate the considerable  improvement of {\em abc-logitboost} over {\em logitboost} and {\em abc-mart}.

\end{abstract}

\section{Introduction}
Boosting\footnote{The idea of {\em abc-logitboost} was included in an unfunded grant proposal submitted in early December 2008.} algorithms \cite{Article:Schapire_ML90,Article:Freund_95,Article:Freund_JCSS97,Article:Bartlett_AS98,Article:Schapire_ML99,Article:FHT_AS00,Proc:Mason_NIPS00,Article:Friedman_AS01} have become very successful in machine learning.  This study revisits {\em logitboost}\cite{Article:FHT_AS00} under the framework of {\em adaptive base class boost (abc-boost)} in \cite{Proc:ABC_ICML09}, for multi-class classification.

We denote a training dataset by $\{y_i,\mathbf{x}_i\}_{i=1}^N$, where $N$ is the number of feature vectors (samples), $\mathbf{x}_i$ is the $i$th feature vector, and  $y_i \in \{0, 1, 2, ..., K-1\}$ is the $i$th class label, where $K\geq 3$ in multi-class classification.

Both {\em logitboost}\cite{Article:FHT_AS00} and {\em mart} (multiple additive regression trees)\cite{Article:Friedman_AS01} algorithms can be viewed as generalizations to the  logistic regression model, which assumes the class probabilities $p_{i,k}$  to be
\begin{align}\label{eqn_logit}
p_{i,k} = \mathbf{Pr}\left(y_i = k|\mathbf{x}_i\right) = \frac{e^{F_{i,k}(\mathbf{x_i})}}{\sum_{s=0}^{K-1} e^{F_{i,s}(\mathbf{x_i})}}, \hspace{0.2in} i = 1, 2, ..., N,
\end{align}
While traditional logistic regression assumes $F_{i,k}(\mathbf{x}_i) = \beta^\text{T}\mathbf{x}_i$, {\em logitboost} and {\em mart} adopt the flexible ``additive model,''  which is a function of $M$ terms:
\begin{align}\label{eqn_F_M}
F^{(M)}(\mathbf{x}) = \sum_{m=1}^M \rho_m h(\mathbf{x};\mathbf{a}_m),
\end{align}
where  $h(\mathbf{x};\mathbf{a}_m)$, the base learner, is typically a regression tree. The parameters, $\rho_m$ and $\mathbf{a}_m$, are learned from the data, by maximum likelihood, which is equivalent to minimizing the {\em negative log-likelihood loss}
\begin{align}\label{eqn_loss}
L = \sum_{i=1}^N L_i, \hspace{0.4in} L_i = - \sum_{k=0}^{K-1}r_{i,k}  \log p_{i,k}
\end{align}
where $r_{i,k} = 1$ if $y_i = k$ and $r_{i,k} = 0$ otherwise.

For identifiability, the ``sum-to-zero'' constraint, $\sum_{k=0}^{K-1}F_{i,k} = 0$, is usually adopted
\cite{Article:FHT_AS00,Article:Friedman_AS01,Article:Zhang_JMLR04,Article:Lee_JASA04,Article:Tewari_JMLR07,Article:Zou_AOAS08}.
\subsection{Logitboost}

As described in Alg. \ref{alg_LogitBoost}, \cite{Article:FHT_AS00} builds the additive model (\ref{eqn_F_M}) by a greedy stage-wise  procedure, using a second-order (diagonal) approximation, which requires knowing the first two derivatives of the loss function (\ref{eqn_loss}) with respective to the function values $F_{i,k}$. \cite{Article:FHT_AS00} obtained:
\begin{align}\label{eqn_mart_d1d2}
&\frac{\partial L_i}{\partial F_{i,k}} = - \left(r_{i,k} - p_{i,k}\right),\hspace{0.2in}
&\frac{\partial^2 L_i}{\partial F_{i,k}^2} = p_{i,k}\left(1-p_{i,k}\right).
\end{align}
Those derivatives can be derived by assuming no relations among $F_{i,k}$, $k = 0$ to $K-1$. However, \cite{Article:FHT_AS00} used the ``sum-to-zero'' constraint $\sum_{k=0}^{K-1}F_{i,k} = 0$ throughout the paper and they provided an alternative explanation.   \cite{Article:FHT_AS00}  showed (\ref{eqn_mart_d1d2}) by conditioning on a ``base class'' and noticed the resultant derivatives are independent of the particular choice of the base class.

{\begin{algorithm}[h]{\small
0: $r_{i,k} = 1$, if $y_{i} = k$, $r_{i,k} =0$ otherwise.\\
1: $F_{i,k} = 0$,\ \  $p_{i,k} = \frac{1}{K}$, \ \ \ $k = 0$ to  $K-1$, \ $i = 1$ to $N$ \\
2: For $m=1$ to $M$ Do\\
3: \hspace{0.2in}    For $k=0$ to $K-1$, Do\\
4: \hspace{0.4in}    Compute $w_{i,k} = p_{i,k}\left(1-p_{i,k}\right)$.\\
5: \hspace{0.4in}    Compute $z_{i,k} = \frac{r_{i,k} - p_{i,k}}{p_{i,k}\left(1-p_{i,k}\right) }$.\\
6: \hspace{0.4in}   Fit the function $f_{i,k}$  by a weighted  least-square of $z_{i,k}$ to $\mathbf{x}_i$ with weights $w_{i,k}$.\\
7:  \hspace{0.4in}  $F_{i,k} = F_{i,k} + \nu \frac{K-1}{K}\left( f_{i,k} - \frac{1}{K}\sum_{k=0}^{K-1}f_{i,k}\right)$\\
8:   \hspace{0.2in} End\\
9: \hspace{0.2in}  $p_{i,k} = \exp(F_{i,k})/\sum_{s=0}^{K-1}\exp(F_{i,s})$, \ \ \ $k = 0$ to  $K-1$, \ $i = 1$ to $N$ \\
10: End
\caption{\small LogitBoost\cite[Alg. 6]{Article:FHT_AS00}. $\nu$ is the shrinkage (e.g., $\nu = 0.1$).\label{alg_LogitBoost} }}

\end{algorithm}}

At each stage, {\em logitboost} fits an individual regression function separately for each class. This is analogous to the popular {\em individualized regression} approach in multinomial logistic regression, which is known \cite{Article:Begg_84,Book:Agresti} to result in  loss of statistical efficiency, compared to the full (conditional) maximum likelihood approach.

On the other hand, in order to use trees as base learner, the diagonal approximation appears to be a must, at least from the practical perspective.

\subsection{Adaptive Base Class Boost}

\cite{Proc:ABC_ICML09} derived the derivatives of (\ref{eqn_loss}) under the sum-to-zero constraint. Without loss of generality, we can assume that class 0 is the base class. For any $k\neq 0$,
\begin{align}\label{eqn_abc_derivatives}
&\frac{\partial L_i}{\partial F_{i,k}}  = \left(r_{i,0} - p_{i,0}\right) - \left(r_{i,k} - p_{i,k}\right),\hspace{0.5in}
\frac{\partial^2 L_i}{\partial F_{i,k}^2} = p_{i,0}(1-p_{i,0}) + p_{i,k}(1-p_{i,k}) + 2p_{i,0}p_{i,k}.
\end{align}
The base class must be identified at each boosting iteration during training. \cite{Proc:ABC_ICML09} suggested an exhaustive procedure to adaptively find the best base class to minimize the training loss (\ref{eqn_loss}) at each iteration.

\cite{Proc:ABC_ICML09} combined the idea of {\em abc-boost} with {\em mart}. The algorithm, {\em abc-mart}, achieved good performance in multi-class classification on the  datasets used in \cite{Proc:ABC_ICML09}.

\subsection{Our Contributions}

We propose {\em abc-logitboost}, by combining {\em abc-boost} with {\em robust logitboost}\cite{Report:Li_Robust-LogitBoost}. Our extensive experiments will demonstrate that {\em abc-logitboost} can considerably improve {\em logitboost} and {\em abc-mart} on a variety of datasets.

\section{Robust Logitboost}

Our work is based on {\em robust logitboost}\cite{Report:Li_Robust-LogitBoost}, which differs from the original {\em logitboost} algorithm. Thus, this section provides an introduction to {\em robust logitboost}.

\cite{Article:Friedman_AS01,Article:FHT_JMLR08} commented that {\em logitboost} (Alg. \ref{alg_LogitBoost}) can be numerically unstable. The original paper\cite{Article:FHT_AS00} suggested some ``crucial implementation protections'' on page 17 of \cite{Article:FHT_AS00}:
\begin{itemize}
\item In Line 5 of Alg. \ref{alg_LogitBoost}, compute the response $z_{i,k}$ by $\frac{1}{p_{i,k}}$ (if $r_{i,k}=1$) or $\frac{-1}{1-p_{i,k}}$ (if $r_{i,k}=0$).
\item Bound the response $|z_{i,k}|$ by $z_{max}\in[2,4]$.
\end{itemize}
Note that the above operations are applied to each individual sample. The goal is to ensure that the response $|z_{i,k}|$ is not too large (Note that $|z_{i,k}|>1$ always). On the other hand, we should hope to use larger $|z_{i,k}|$ to better capture the data variation. Therefore, the thresholding occurs very frequently and it is expected that some of the useful information is lost.

 \cite{Report:Li_Robust-LogitBoost} demonstrated that, if implemented carefully, {\em logitboost} is almost identical to {\em mart}. The only difference is the tree-splitting criterion.

\subsection{The Tree-Splitting Criterion Using the Second-Order Information}\label{sec_split}

Consider $N$ weights $w_i$, and $N$ response values $z_i$, $i=1$ to $N$, which are assumed to be ordered according to the sorted order of the corresponding feature values. The tree-splitting procedure is to find the index $s$, $1\leq s<N$, such that the weighted mean square error (MSE) is reduced the most if split at $s$.  That is, we seek $s$ to maximize
{\small\begin{align}\notag
Gain(s) =& MSE_{T} - (MSE_{L} + MSE_{R})=\sum_{i=1}^N (z_i - \bar{z})^2w_i - \left[
\sum_{i=1}^s (z_i - \bar{z}_L)^2w_i + \sum_{i=s+1}^N (z_i - \bar{z}_R)^2w_i\right]
\end{align}}
where $\bar{z} = \frac{\sum_{i=1}^N z_iw_i}{\sum_{i=1}^N w_i}$,
$\bar{z}_L = \frac{\sum_{i=1}^s z_iw_i}{\sum_{i=1}^s w_i}$, and
$\bar{z}_R = \frac{\sum_{i=s+1}^N z_iw_i}{\sum_{i=s+1}^{N} w_i}$. After simplification, we obtain
\begin{align}\notag
Gain(s) = \frac{\left[\sum_{i=1}^s z_iw_i\right]^2}{\sum_{i=1}^s w_i}+\frac{\left[\sum_{i=s+1}^N z_iw_i\right]^2}{\sum_{i=s+1}^{N} w_i} - \frac{\left[\sum_{i=1}^N z_iw_i\right]^2}{\sum_{i=1}^N w_i}
\end{align}
Plugging in  $w_i = p_{i,k}(1-p_{i,k})$, and $z_i = \frac{r_{i,k}-p_{i,k}}{p_{i,k}(1-p_{i,k})}$ as in Alg. \ref{alg_LogitBoost}, yields,
\begin{align}\notag
Gain(s) =  \frac{\left[\sum_{i=1}^s r_{i,k} - p_{i,k} \right]^2}{\sum_{i=1}^s p_{i,k}(1-p_{i,k})}+\frac{\left[\sum_{i=s+1}^N r_{i,k} - p_{i,k} \right]^2}{\sum_{i=s+1}^{N} p_{i,k}(1-p_{i,k})} - \frac{\left[\sum_{i=1}^N r_{i,k} - p_{i,k} \right]^2}{\sum_{i=1}^N p_{i,k}(1-p_{i,k})}.
\end{align}
Because the computations involve $\sum p_{i,k}(1-p_{i,k})$ as a group, this procedure is actually numerically stable.\\

In comparison, {\em mart}\cite{Article:Friedman_AS01} only used the first order information to construct the trees, i.e.,
\begin{align}\notag
MART Gain(s) =  \left[\sum_{i=1}^s r_{i,k} - p_{i,k} \right]^2+\left[\sum_{i=s+1}^N r_{i,k} - p_{i,k} \right]^2 - \left[\sum_{i=1}^N r_{i,k} - p_{i,k} \right]^2.
\end{align}

\subsection{The Robust Logitboost Algorithm}

{\scriptsize\begin{algorithm}{\small
1: $F_{i,k} = 0$, $p_{i,k} = \frac{1}{K}$, $k = 0$ to  $K-1$, $i = 1$ to $N$ \\
2: For $m=1$ to $M$ Do\\
3: \hspace{0.1in}    For $k=0$ to $K-1$ Do\\
%4: \hspace{0.2in}  $p_{i,k} = \exp(F_{i,k})/\sum_{s=0}^{K-1}\exp(F_{i,s})$\\
4:  \hspace{0.2in}  $\left\{R_{j,k,m}\right\}_{j=1}^J = J$-terminal node regression tree from $\{r_{i,k} - p_{i,k}, \ \ \mathbf{x}_{i}\}_{i=1}^N$,\\\notag
: \hspace{1.0in} with weights $p_{i,k}(1-p_{i,k})$ as in Section 2.1. \\
5:   \hspace{0.2in}  $\beta_{j,k,m} = \frac{K-1}{K}\frac{ \sum_{\mathbf{x}_i \in
  R_{j,k,m}} r_{i,k} - p_{i,k}}{ \sum_{\mathbf{x}_i\in
  R_{j,k,m}}\left(1-p_{i,k}\right)p_{i,k} }$ \\
6:  \hspace{0.2in}  $F_{i,k} = F_{i,k} +
\nu\sum_{j=1}^J\beta_{j,k,m}1_{\mathbf{x}_i\in R_{j,k,m}}$ \\
7:   \hspace{0.1in} End\\
8:\hspace{0.12in} $p_{i,k} = \exp(F_{i,k})/\sum_{s=0}^{K-1}\exp(F_{i,s})$, \ \ \ $k = 0$ to  $K-1$, \ $i = 1$ to $N$ \\
9: End
\caption{\small Robust logitboost, which is very similar to mart, except for Line 4.  }
\label{alg_robust_logitboost}}
\end{algorithm}}

Alg. \ref{alg_robust_logitboost} describes {\em robust logitboost} using the tree-splitting criterion developed  in Section \ref{sec_split}. Note that after trees are constructed, the values of the terminal nodes are computed by
\begin{align}\notag
\frac{\sum_{node} z_{i,k} w_{i,k}}{\sum_{node} w_{i,k}} = \frac{\sum_{node} r_{i,k} - p_{i,k}}{\sum_{node} p_{i,k}(1-p_{i,k})},
\end{align}
which explains Line 5 of Alg. \ref{alg_robust_logitboost}.

\subsubsection{Three Main Parameters: $J$, $\nu$, and $M$}

Alg. \ref{alg_robust_logitboost} has three main parameters, to which the performance is not very sensitive, as long as they fall in some reasonable range. This is a very significant advantage in practice.

The number of terminal nodes, $J$,  determines the capacity of the base learner.  \cite{Article:Friedman_AS01} suggested $J=6$. \cite{Article:FHT_AS00,Article:Zou_AOAS08} commented that $J> 10$ is unlikely. In our experience, for large datasets (or moderate datasets in high-dimensions), $J=20$ is often a reasonable choice; also see \cite{Proc:McRank_NIPS07}.

The shrinkage, $\nu$, should be large enough to make sufficient progress at each step and  small enough to avoid over-fitting.  \cite{Article:Friedman_AS01} suggested $\nu\leq 0.1$. Normally, $\nu=0.1$ is  used.

The number of iterations, $M$, is largely determined  by the affordable computing time. A commonly-regarded merit of boosting is that over-fitting can be largely avoided for reasonable $J$ and $\nu$.

\section{Adaptive Base Class Logitboost}

\begin{algorithm}[h]{\small
1: $F_{i,k} = 0$,\ \  $p_{i,k} = \frac{1}{K}$, \ \ \ $k = 0$ to  $K-1$, \ $i = 1$ to $N$ \\
2: For $m=1$ to $M$ Do\\
%3: \hspace{0.1in}  $G_{i,k,b} = F_{i,k}$\\
3: \hspace{0.1in}    For $b=0$ to $K-1$, Do\\
4: \hspace{0.2in}    For $k=0$ to $K-1$, $k\neq b$, Do\\
5:  \hspace{0.3in}  $\left\{R_{j,k,m}\right\}_{j=1}^J = J$-terminal
node regression tree from  $\{-(r_{i,b} - p_{i,b}) +  (r_{i,k} - p_{i,k}), \ \ \mathbf{x}_{i}\}_{i=1}^N$\\
: \hspace{1.0in} with weights $p_{i,b}(1-p_{i,b})+p_{i,k}(1-p_{i,k})+2p_{i,b}p_{i,k}$, as in Section 2.1.
 \\
6:   \hspace{0.3in}  $\beta_{j,k,m} = \frac{ \sum_{\mathbf{x}_i \in
  R_{j,k,m}} -(r_{i,b} - p_{i,b}) + (r_{i,k} - p_{i,k})  }{ \sum_{\mathbf{x}_i\in
  R_{j,k,m}} p_{i,b}(1-p_{i,b})+ p_{i,k}\left(1-p_{i,k}\right) + 2p_{i,b}p_{i,k} }$ \\\\
7:  \hspace{0.3in}  $G_{i,k,b} = F_{i,k} +
\nu\sum_{j=1}^J\beta_{j,k,m}1_{\mathbf{x}_i\in R_{j,k,m}}$ \\
8:   \hspace{0.2in} End\\
9: \hspace{0.2in} $G_{i,b,b} = - \sum_{k\neq b} G_{i,k,b}$ \\
10: \hspace{0.2in}  $q_{i,k} = \exp(G_{i,k,b})/\sum_{s=0}^{K-1}\exp(G_{i,s,b})$ \\
11: \hspace{0.2in} $L^{(b)} = -\sum_{i=1}^N \sum_{k=0}^{K-1} r_{i,k}\log\left(q_{i,k}\right)$\\
12: \hspace{0.1in} End\\
13: \hspace{0.1in} $B(m) = \underset{b}{\text{argmin}} \  \ L^{(b)}$\\
14: \hspace{0.1in} $F_{i,k} = G_{i,k,B(m)}$\\
15:\hspace{0.1in}  $p_{i,k} = \exp(F_{i,k})/\sum_{s=0}^{K-1}\exp(F_{i,s})$ \\
16: End}
\caption{{\em Abc-logitboost} using the exhaustive search strategy for the base class, as suggested in \cite{Proc:ABC_ICML09}.  The vector $B$ stores the base class numbers. }
\label{alg_abc-logitboost}
\end{algorithm}%

The recently proposed {\em abc-boost} \cite{Proc:ABC_ICML09} algorithm consists of two key components:
\begin{enumerate}
\item Using the widely-used {\em sum-to-zero} constraint\cite{Article:FHT_AS00,Article:Friedman_AS01,Article:Zhang_JMLR04,Article:Lee_JASA04,Article:Tewari_JMLR07,Article:Zou_AOAS08} on the loss function,  one can formulate boosting algorithms only for $K-1$ classes, by using one class as the \textbf{base class}.
\item At each boosting iteration, \textbf{adaptively} select the base class according to the training loss. \cite{Proc:ABC_ICML09} suggested an exhaustive search strategy.
\end{enumerate}

\cite{Proc:ABC_ICML09} combined {\em abc-boost} with {\em mart} to develop {\em abc-mart}. \cite{Proc:ABC_ICML09}  demonstrated the good performance of {\em abc-mart} compared to {\em mart}. This study will illustrate that \textbf{\em abc-logitboost}, the combination of {\em abc-boost} with {\em (robust) logitboost}, will further reduce the test errors, at least on a variety of  datasets.

Alg. \ref{alg_abc-logitboost} presents  {\em abc-logitboost}, using the derivatives in (\ref{eqn_abc_derivatives}) and the same exhaustive search strategy as in {\em abc-mart}. Again, {\em abc-logitboost} differs from {\em abc-mart} only in the tree-splitting procedure (Line 5 in Alg. \ref{alg_abc-logitboost}).

\section{Experiments}\label{sec_exp}

Table \ref{tab_data} lists the datasets in our experiments, which include all the datasets used in \cite{Proc:ABC_ICML09}, plus {\em Mnist10k}\footnote{We also did limited experiments on the original {\em Mnist} dataset (i.e., 60000 training samples and 10000 testing samples), the test mis-classification error rate was about $1.3\%$.}.

\begin{table}[h]
\caption{For {\em Letter, Pendigits, Zipcode, Optdigits} and {\em Isolet}, we used the standard (default) training and test sets. For {\em Covertype}, we use the same split in \cite{Proc:ABC_ICML09}. For {\em Mnist10k}, we used the original 10000 test samples in the original {\em Mnist} dataset for training, and the original 60000 training samples for testing. Also, as explained in \cite{Proc:ABC_ICML09}, {\em Letter2k} ({\em Letter4k}) used the last 2000 (4000) samples of {\em Letter} for training and the remaining 18000 (16000) for testing, from the original {\em Letter} dataset.
 }
\begin{center}{\small
\begin{tabular}{l r r r r}
\hline \hline
dataset &$K$ & \# training & \# test &\# features\\
\hline
Covertype &7 & 290506 & 290506 & 54\\
Mnist10k &10 &10000 &60000&784\\
Letter2k &26   & 2000   &18000 &16\\
Letter4k &26   & 4000   &16000 &16\\
Letter &26   & 16000   &4000 &16\\
Pendigits &10 &7494   &3498 &16\\
Zipcode  &10  &7291   &2007 &256\\
Optdigits &10  &3823   &1797 &64\\
Isolet &26   & 6218   &1559 &617\\
\hline\hline
\end{tabular}
}
\end{center}
\label{tab_data}
\end{table}

Note that {\em Zipcode, Otpdigits,} and {\em Isolet} are very small datasets (especially the testing sets). They may not necessarily provide a reliable comparison of different algorithms. Since they are popular datasets, we nevertheless include them in our experiments.

Recall {\em logitboost} has three main parameters, $J$, $\nu$, and $M$. Since overfitting is largely avoided, we simply let $M=10000$ ($M=5000$ only for {\em Covertype}), unless the machine zero is reached. The performance is not sensitive to $\nu$ (as long as $\nu\leq 0.1$). The performance is also not too sensitive to $J$ in a good range.

Ideally, we would like to show that, for every reasonable combination of $J$ and $\nu$ (using $M$ as large as possible), {\em abc-logitboost} exhibits consistent improvement over {\em (robust) logitboost}. For most datasets, we experimented with every combination of $J\in\{4, 6, 8, 10, 12, 14, 16,18, 20\}$ and $\nu \in\{0.04, 0.06, 0.08, 0.1\}$.

We provide a summary of the experiments after presenting the detailed results on {\em Mnist10k}.

\subsection{Experiments on the {\em Mnist10k} Dataset}
%\vspace{-0.05in}

For this  dataset, we experimented with every combination of $J\in\{4, 6, 8, 10, 12, 14, 16,18, 20\}$ and\\ $\nu \in\{0.04, 0.06, 0.08, 0.1\}$. We trained till the loss (\ref{eqn_loss}) reached the machine zero, to exhaust the capacity of the learner so that we could provide a reliable comparison, up to $M=10000$ iterations.

\begin{figure}[h]
\begin{center}
\mbox{\includegraphics[width=1.6in]{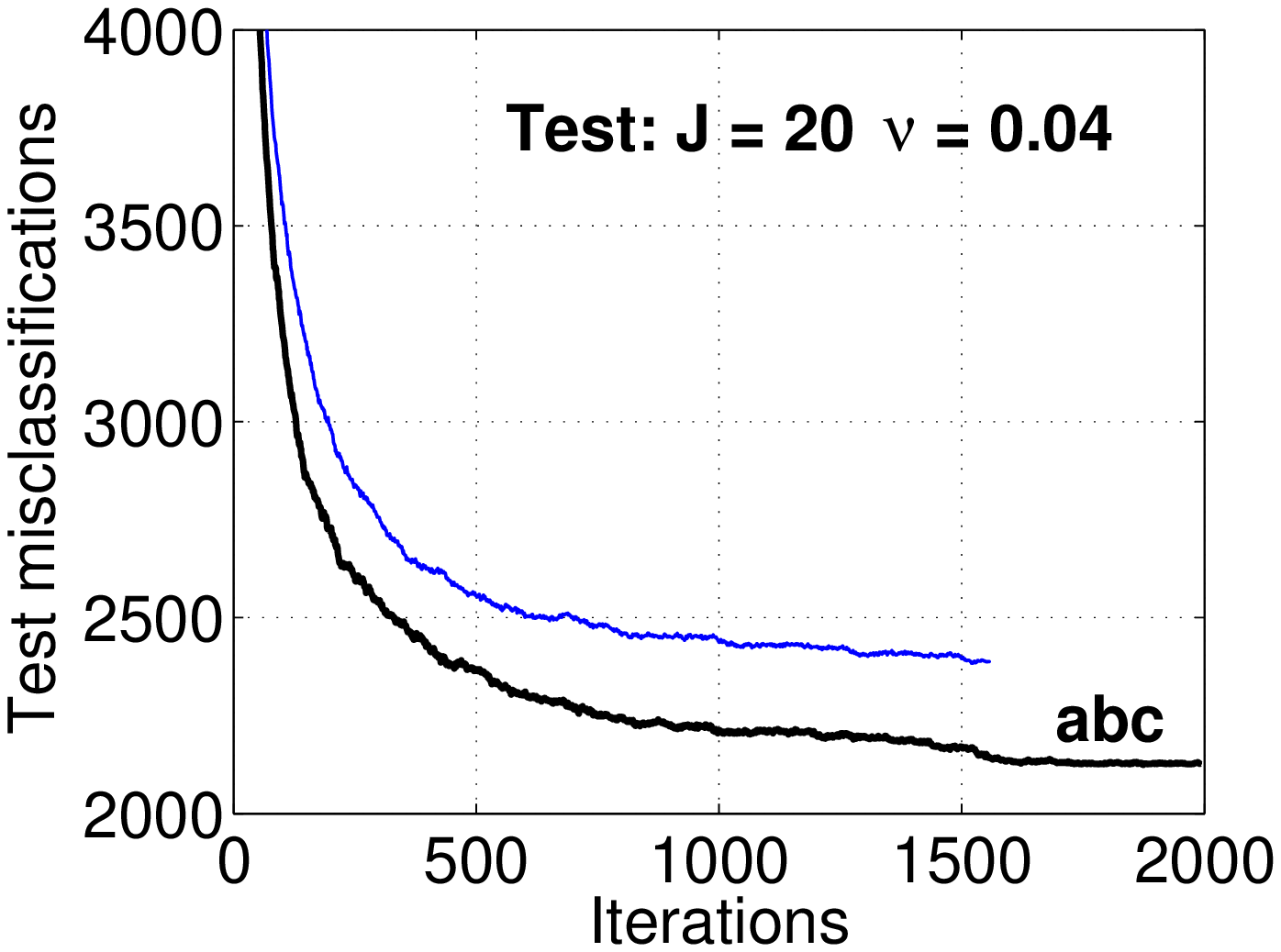}
\includegraphics[width=1.6in]{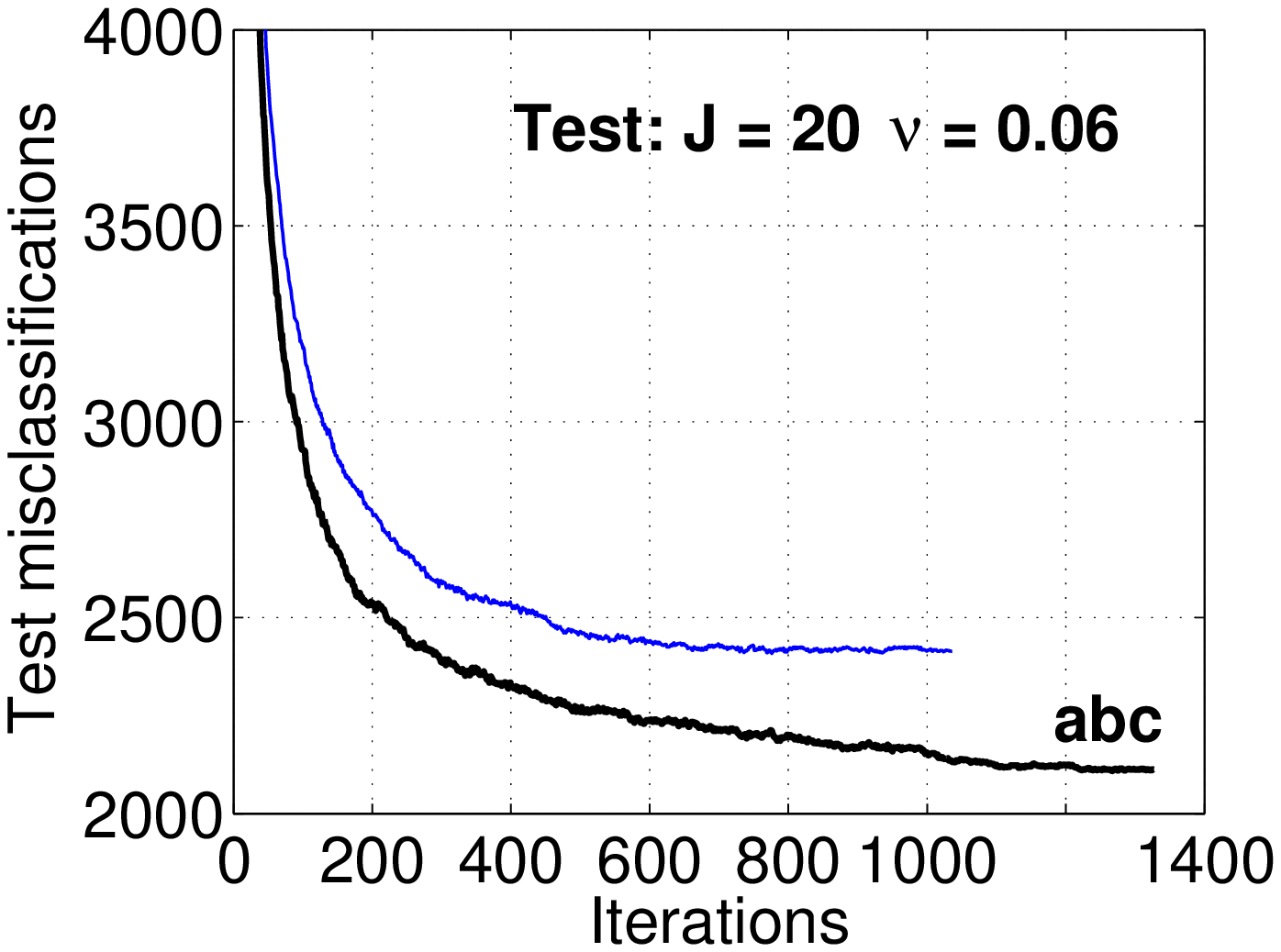}
\includegraphics[width=1.6in]{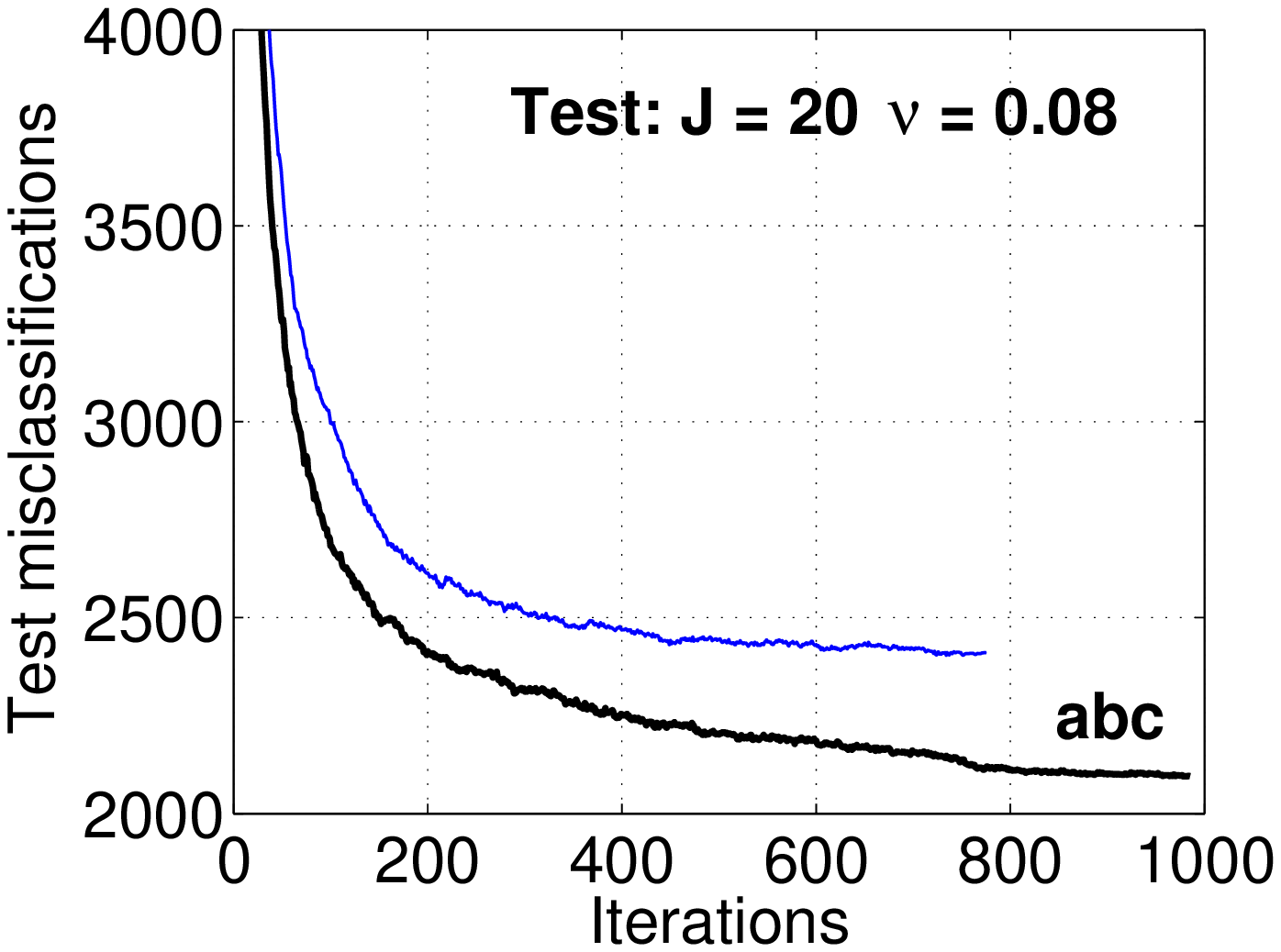}
\includegraphics[width=1.6in]{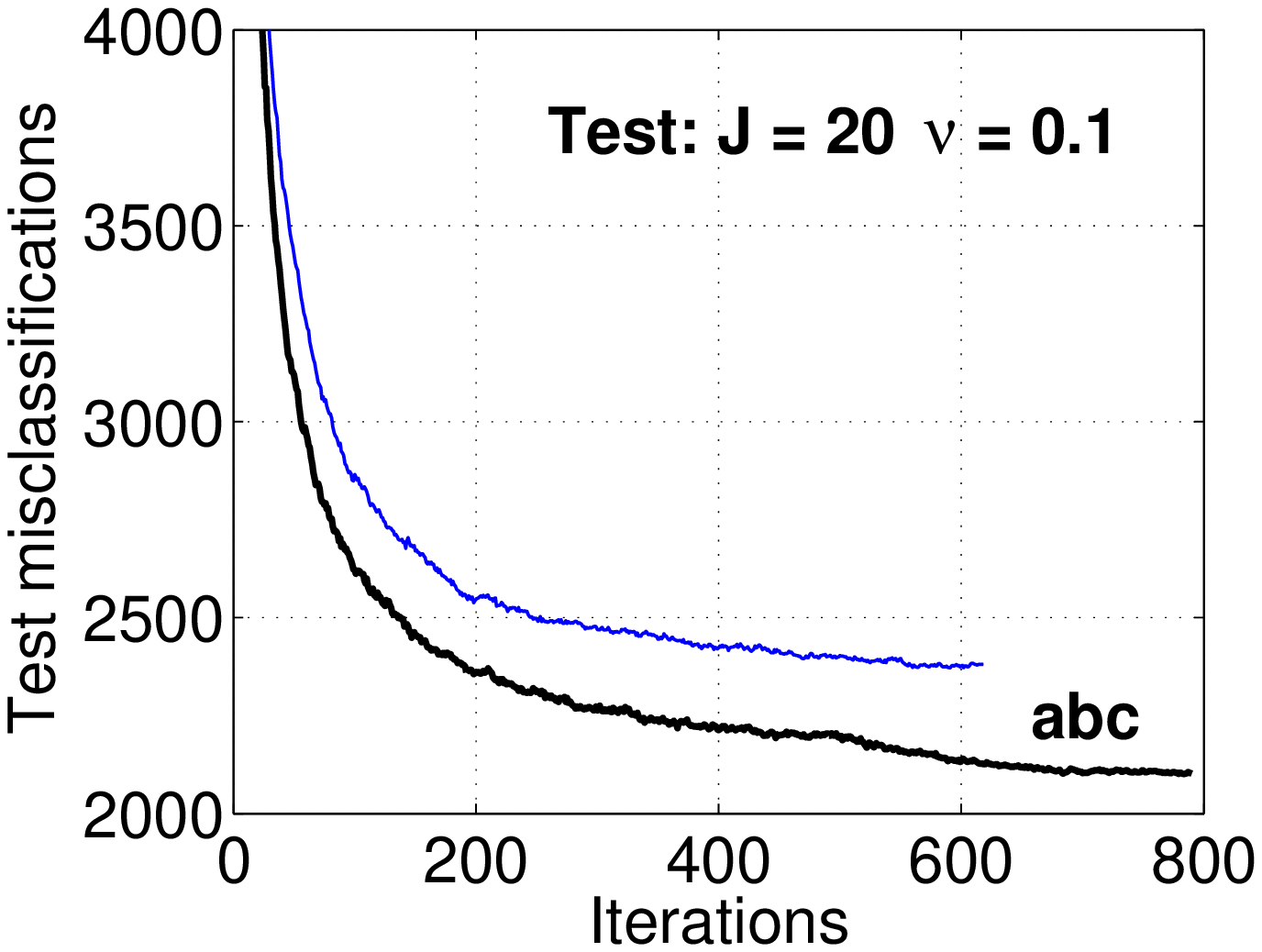}}

\mbox{\includegraphics[width=1.6in]{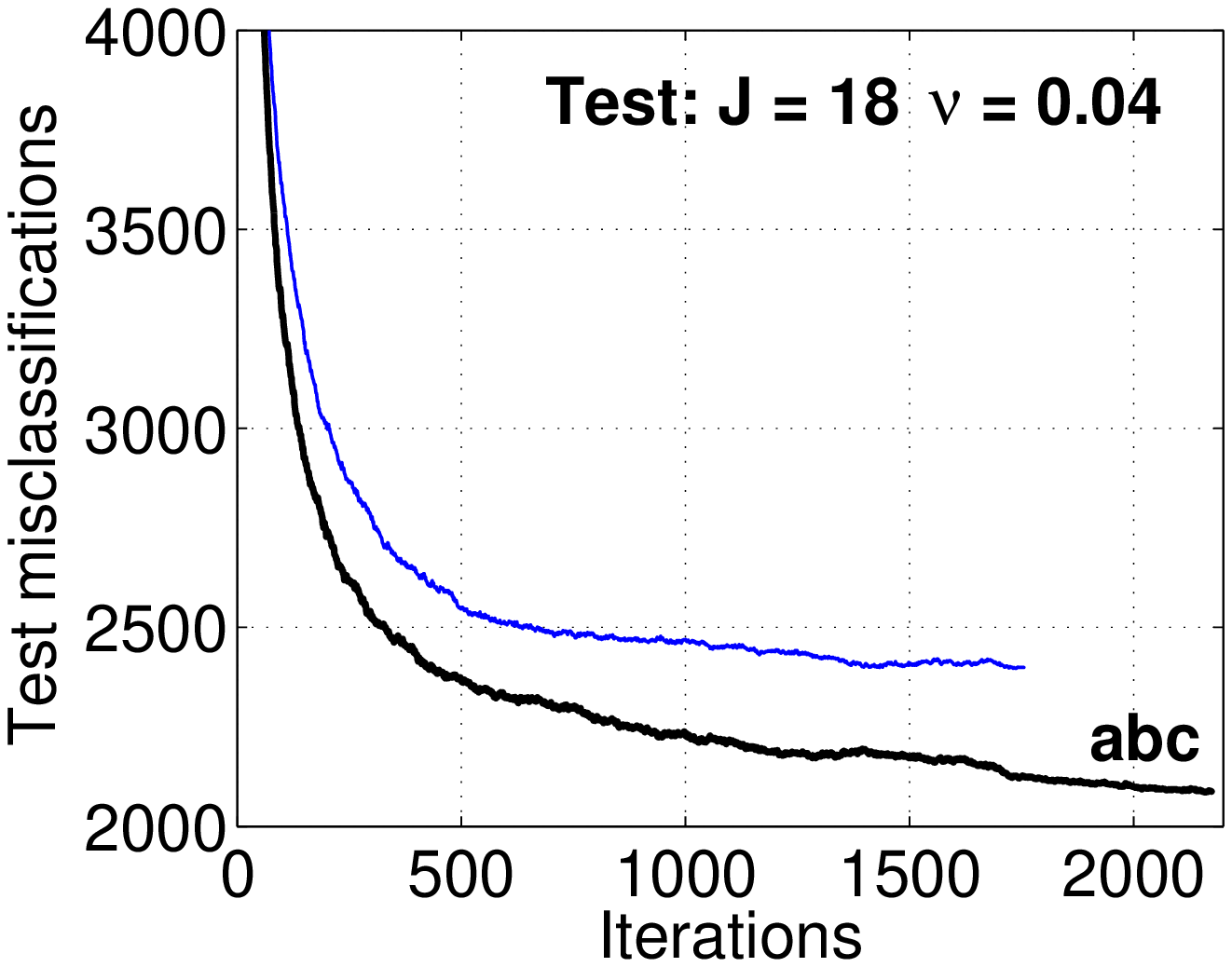}
\includegraphics[width=1.6in]{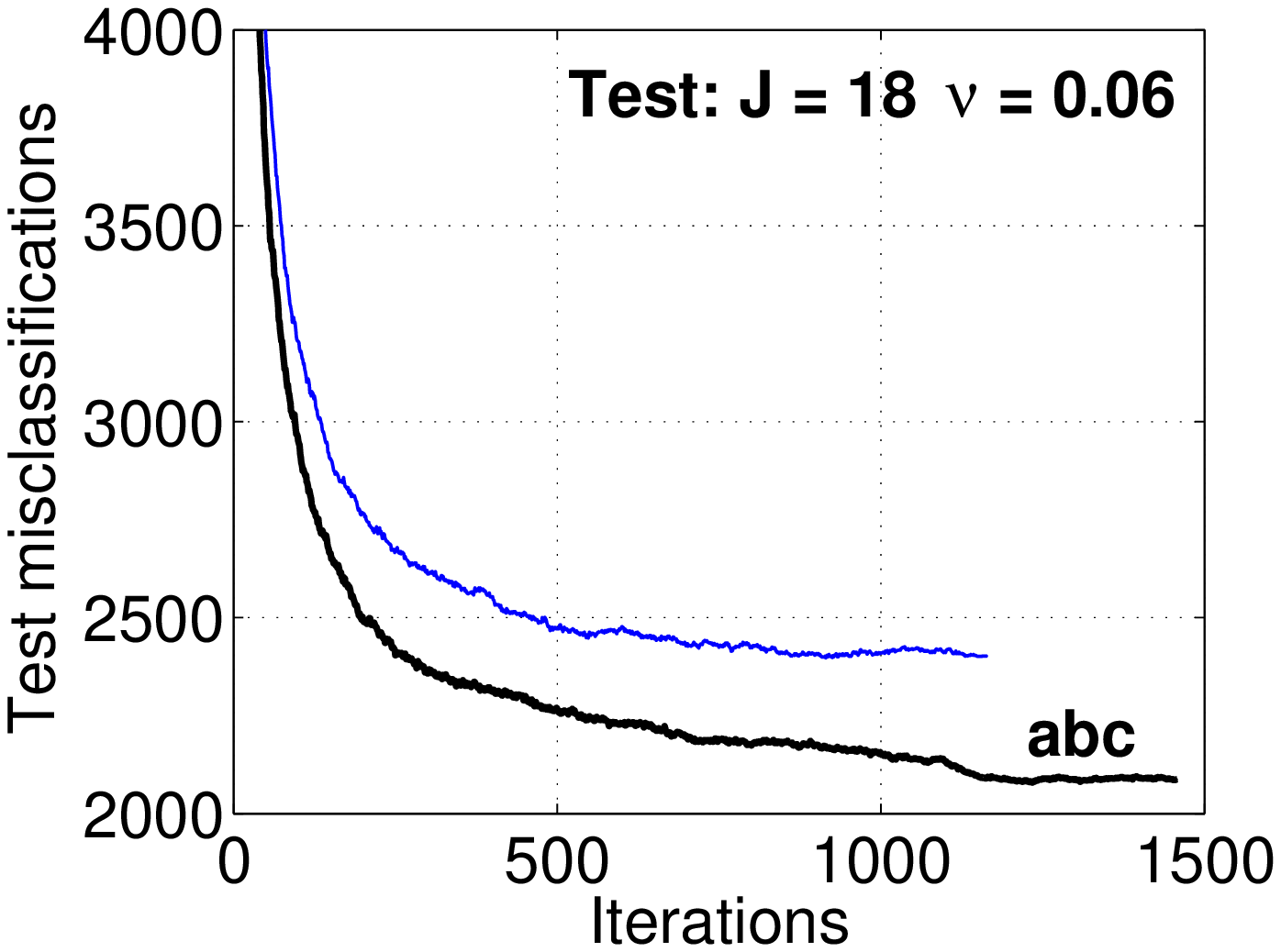}
\includegraphics[width=1.6in]{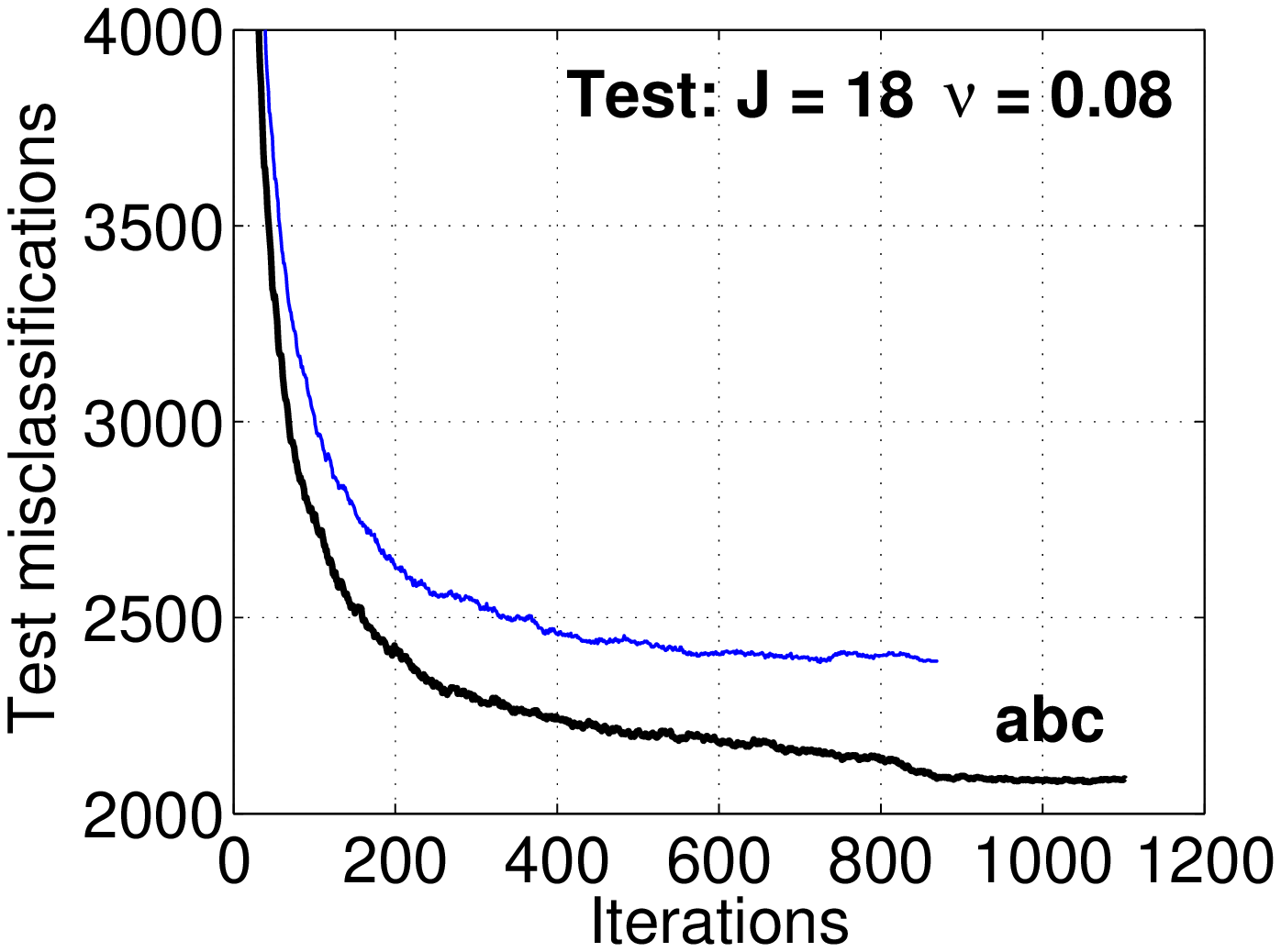}
\includegraphics[width=1.6in]{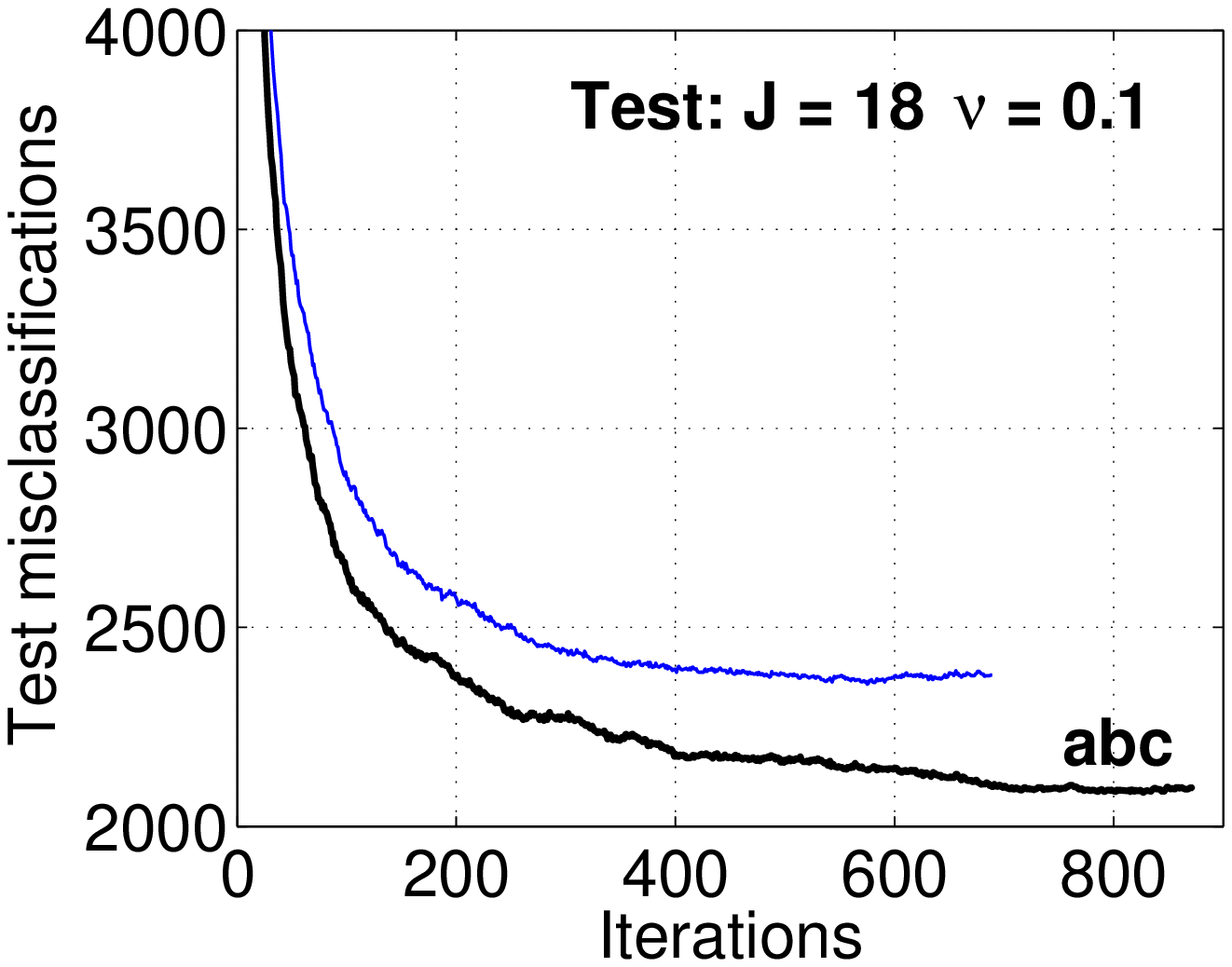}}

\mbox{\includegraphics[width=1.6in]{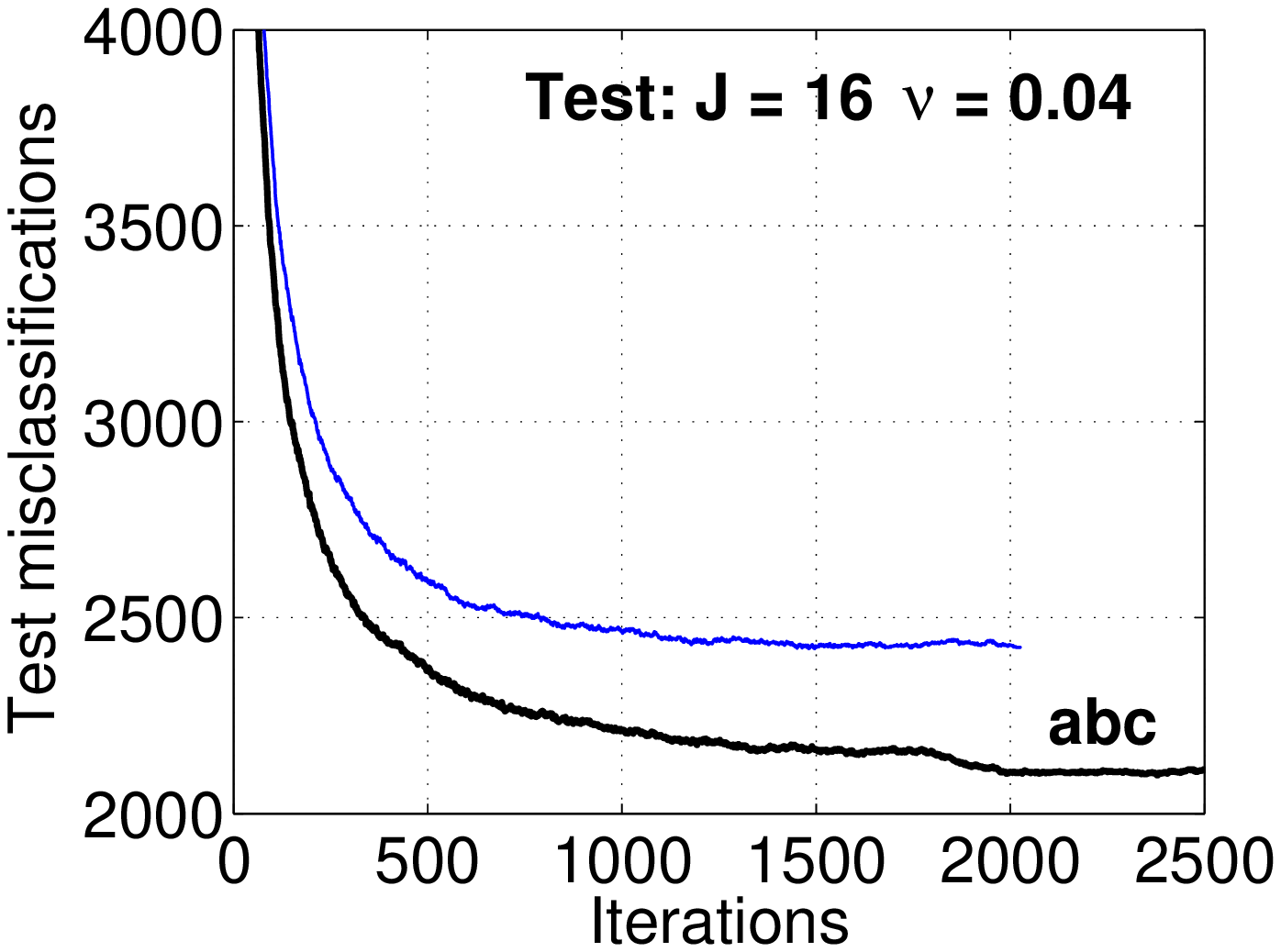}
\includegraphics[width=1.6in]{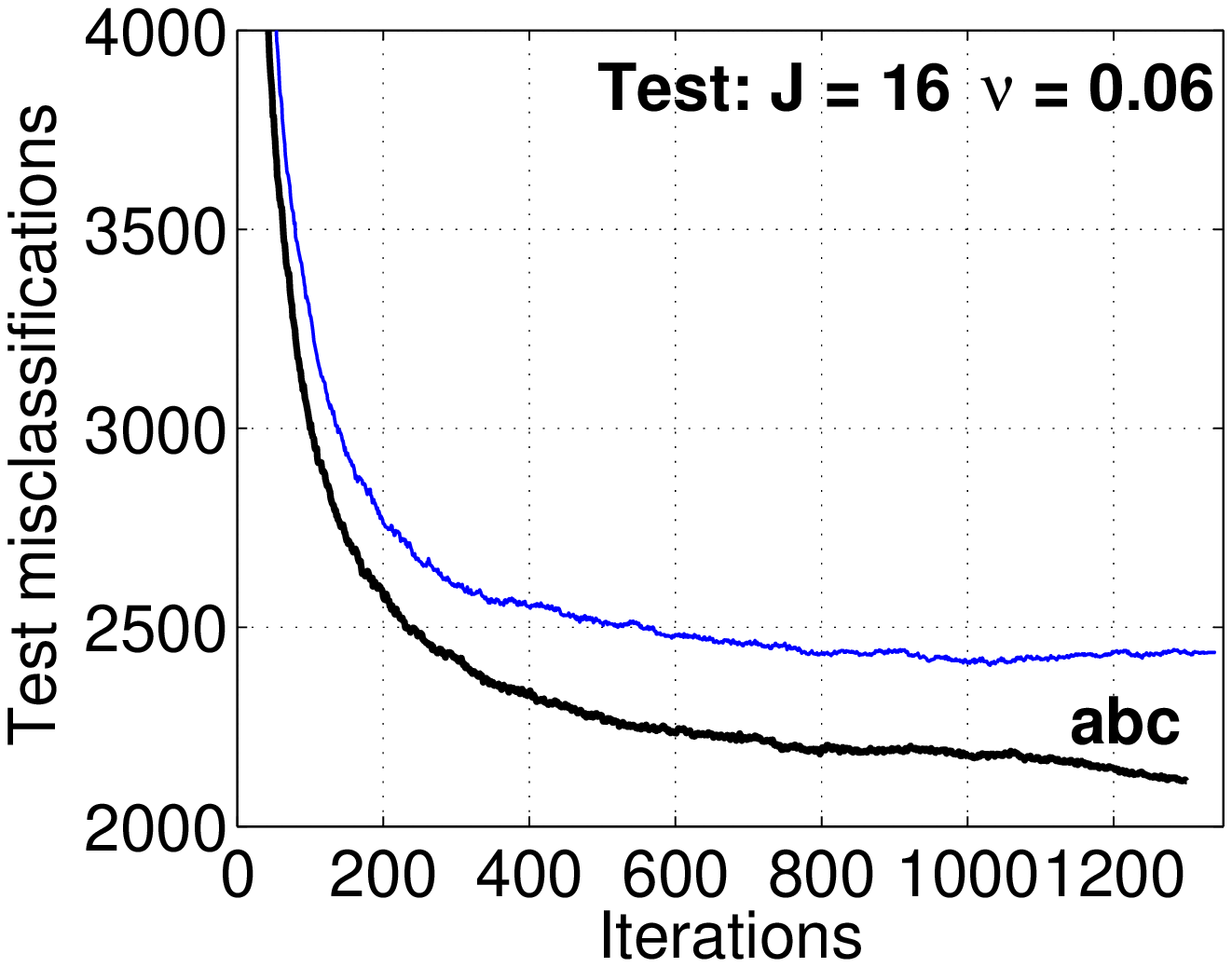}
\includegraphics[width=1.6in]{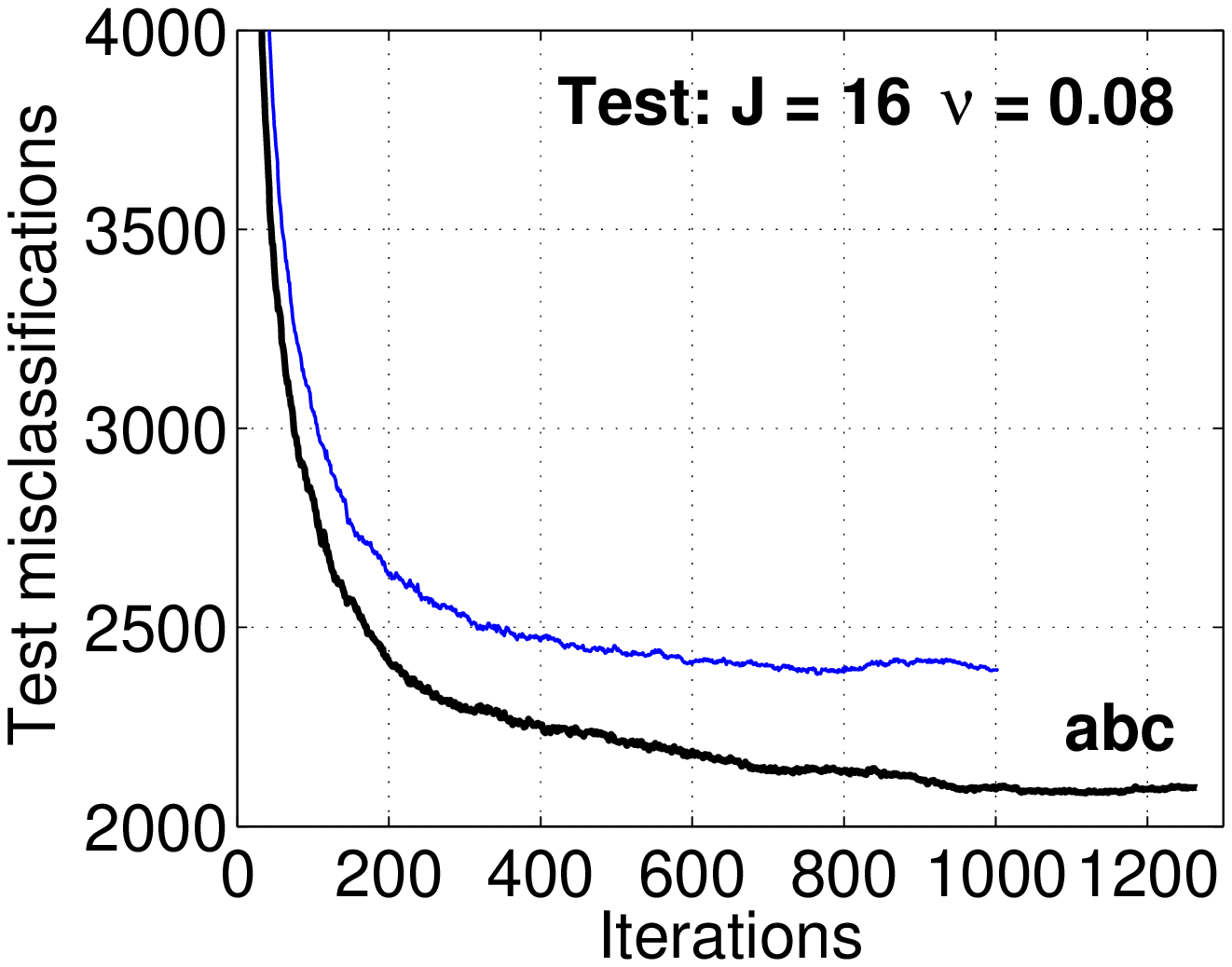}
\includegraphics[width=1.6in]{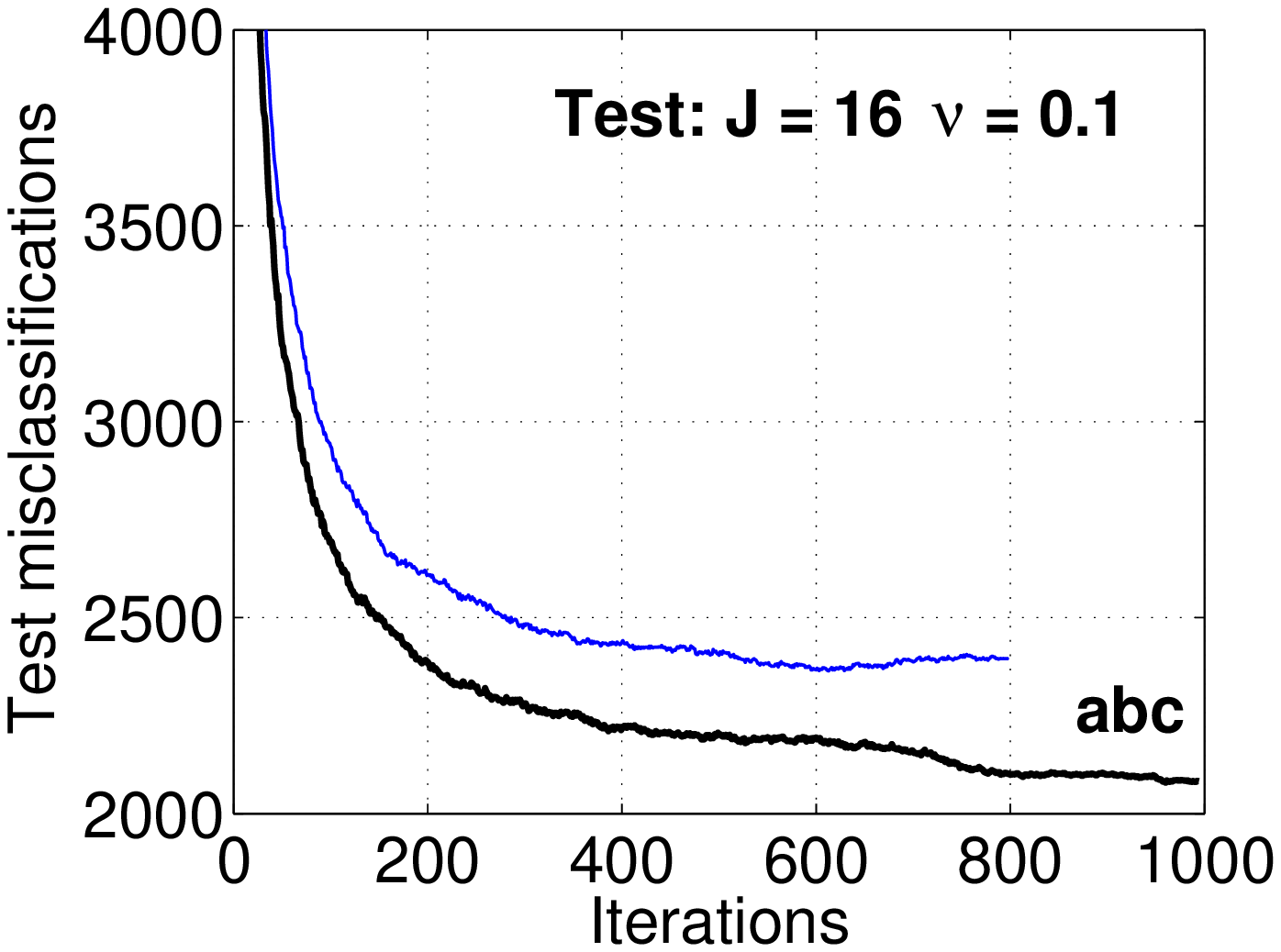}}

\mbox{\includegraphics[width=1.6in]{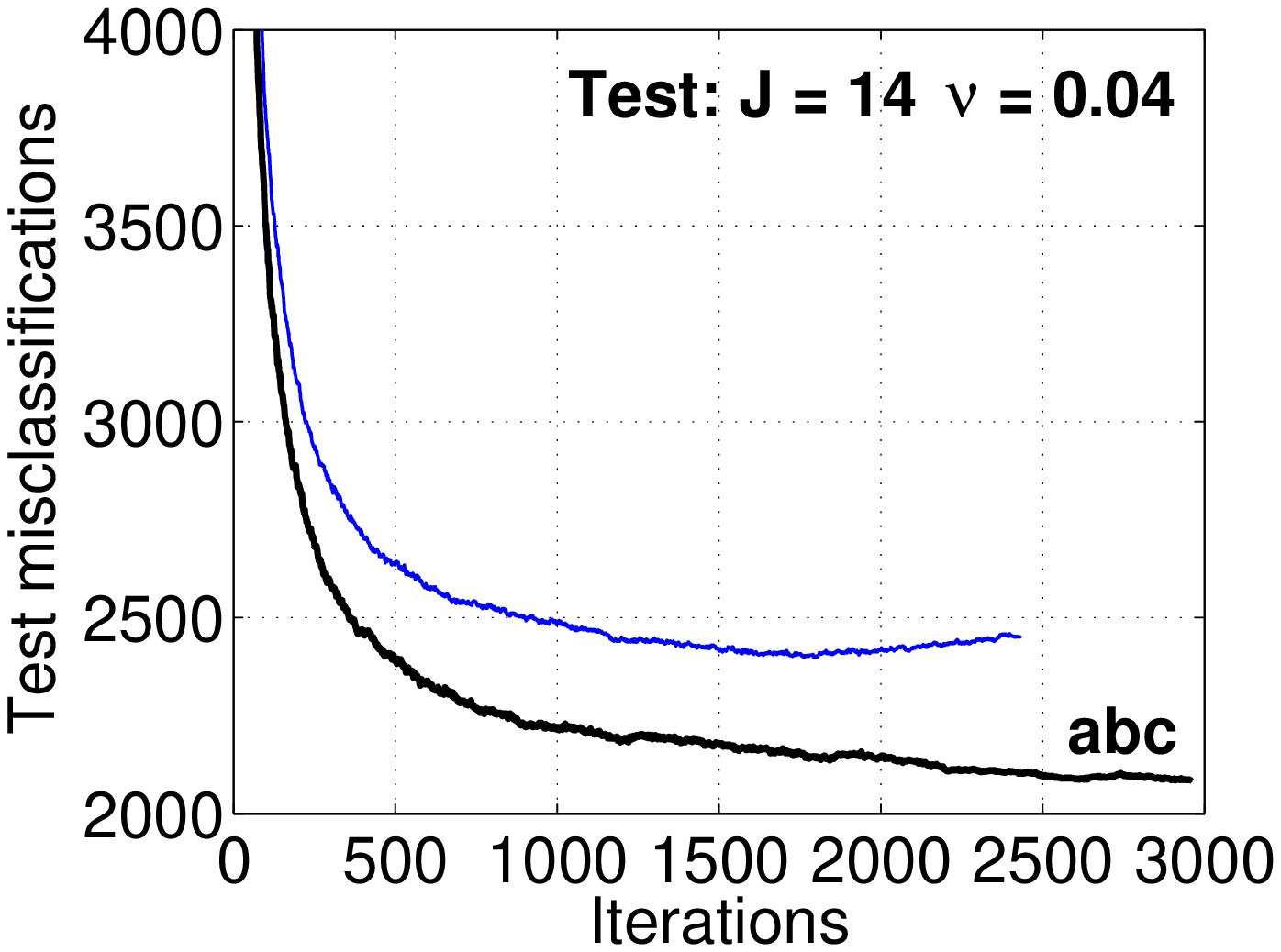}
\includegraphics[width=1.6in]{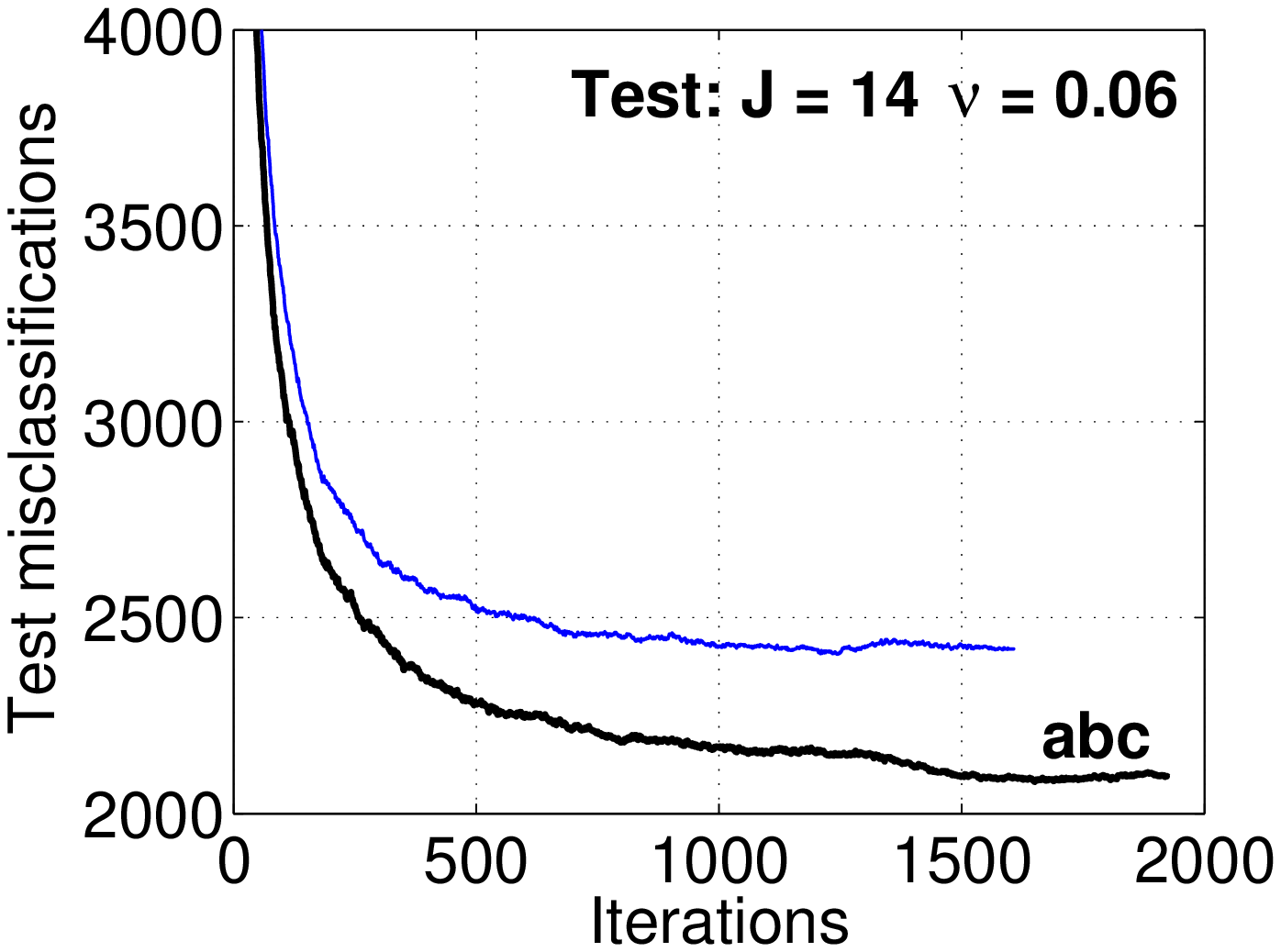}
\includegraphics[width=1.6in]{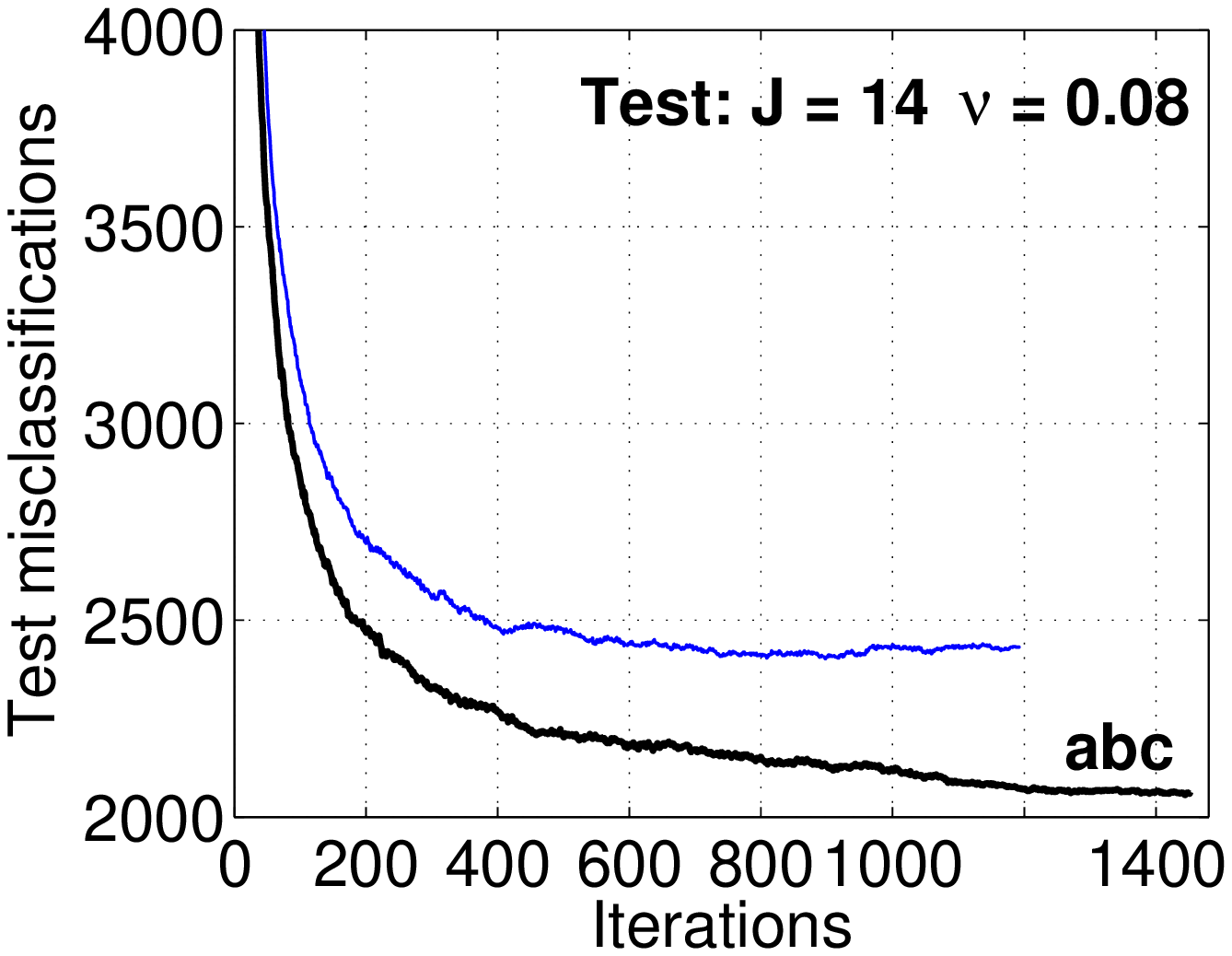}
\includegraphics[width=1.6in]{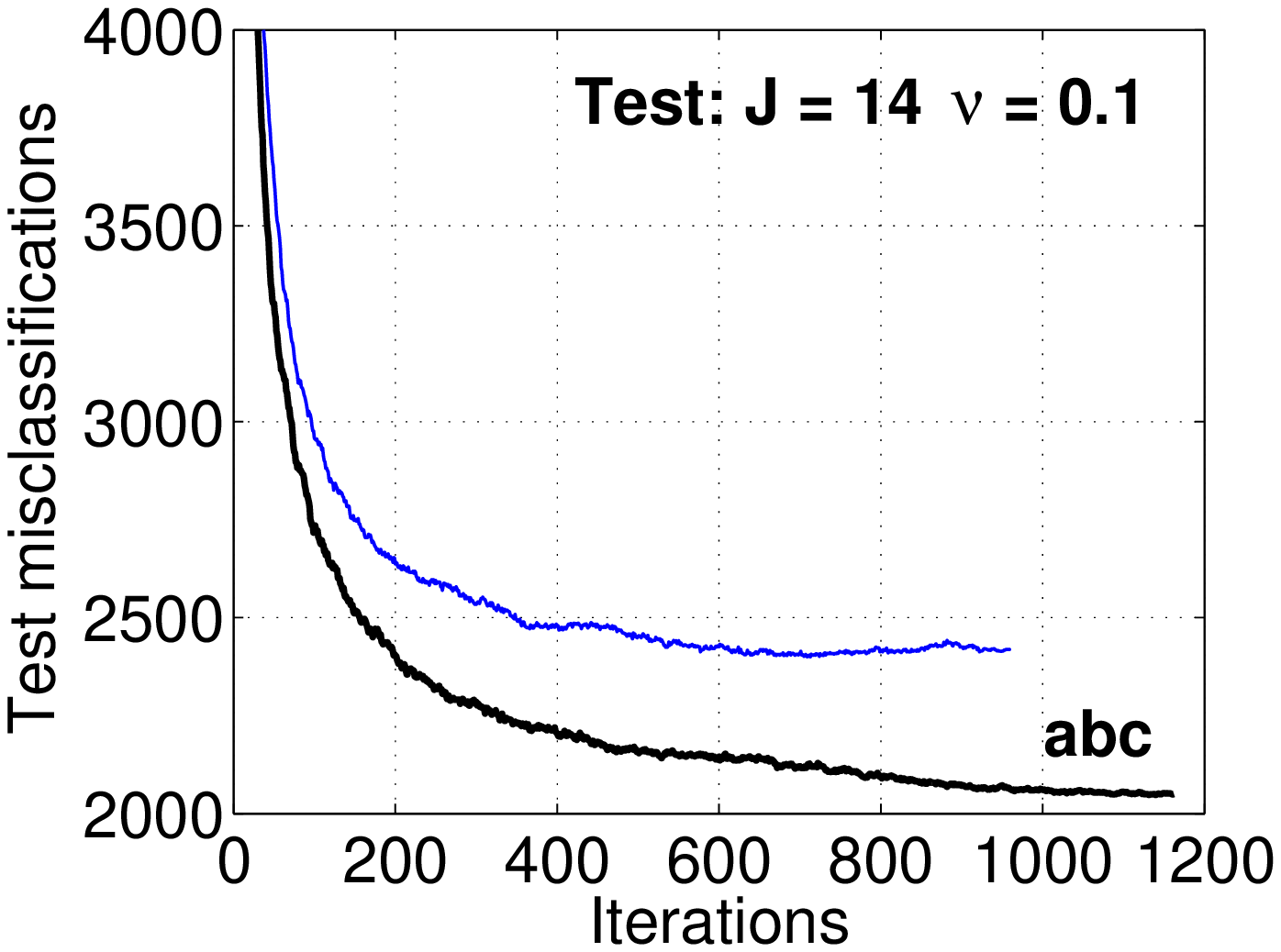}}

\mbox{\includegraphics[width=1.6in]{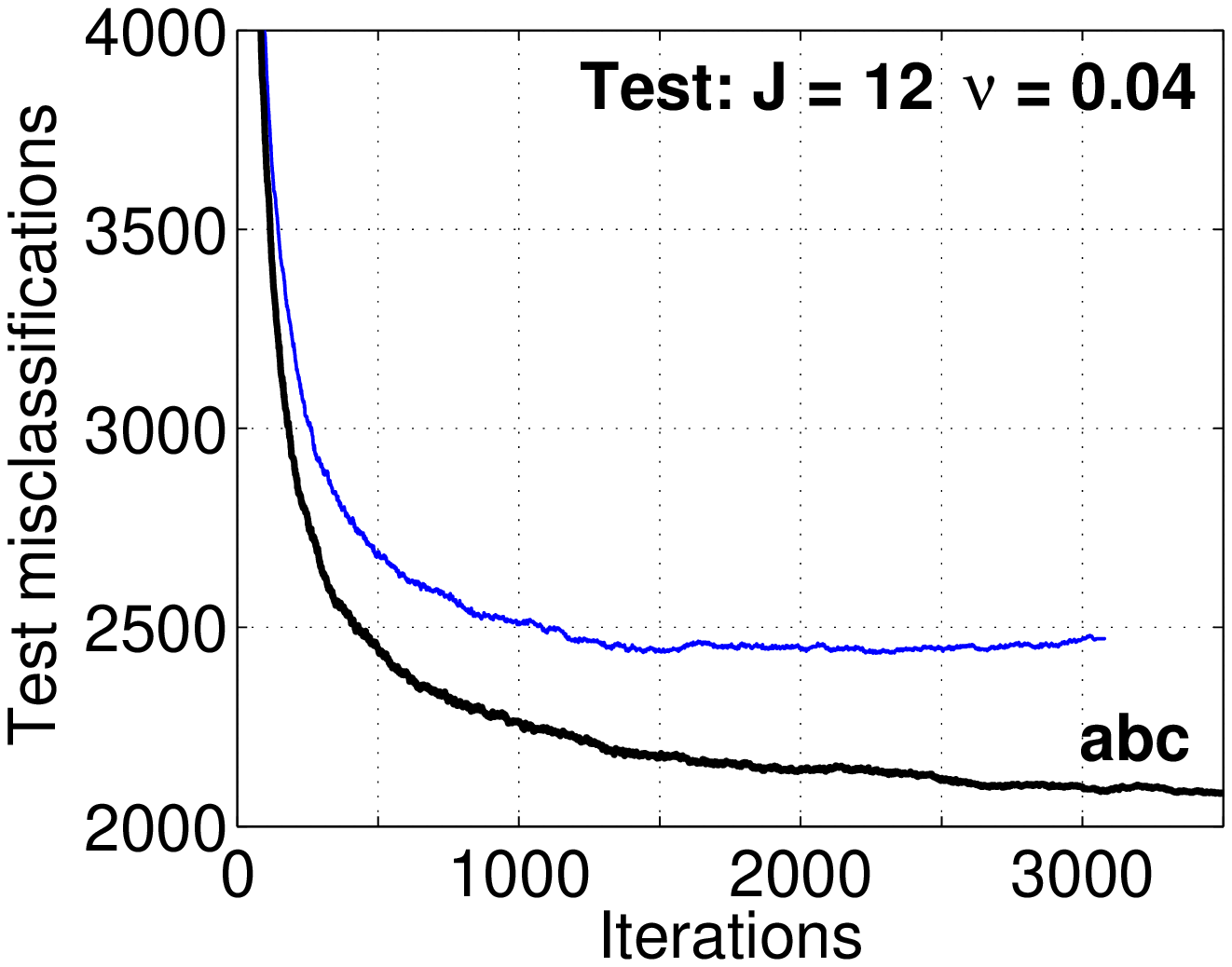}
\includegraphics[width=1.6in]{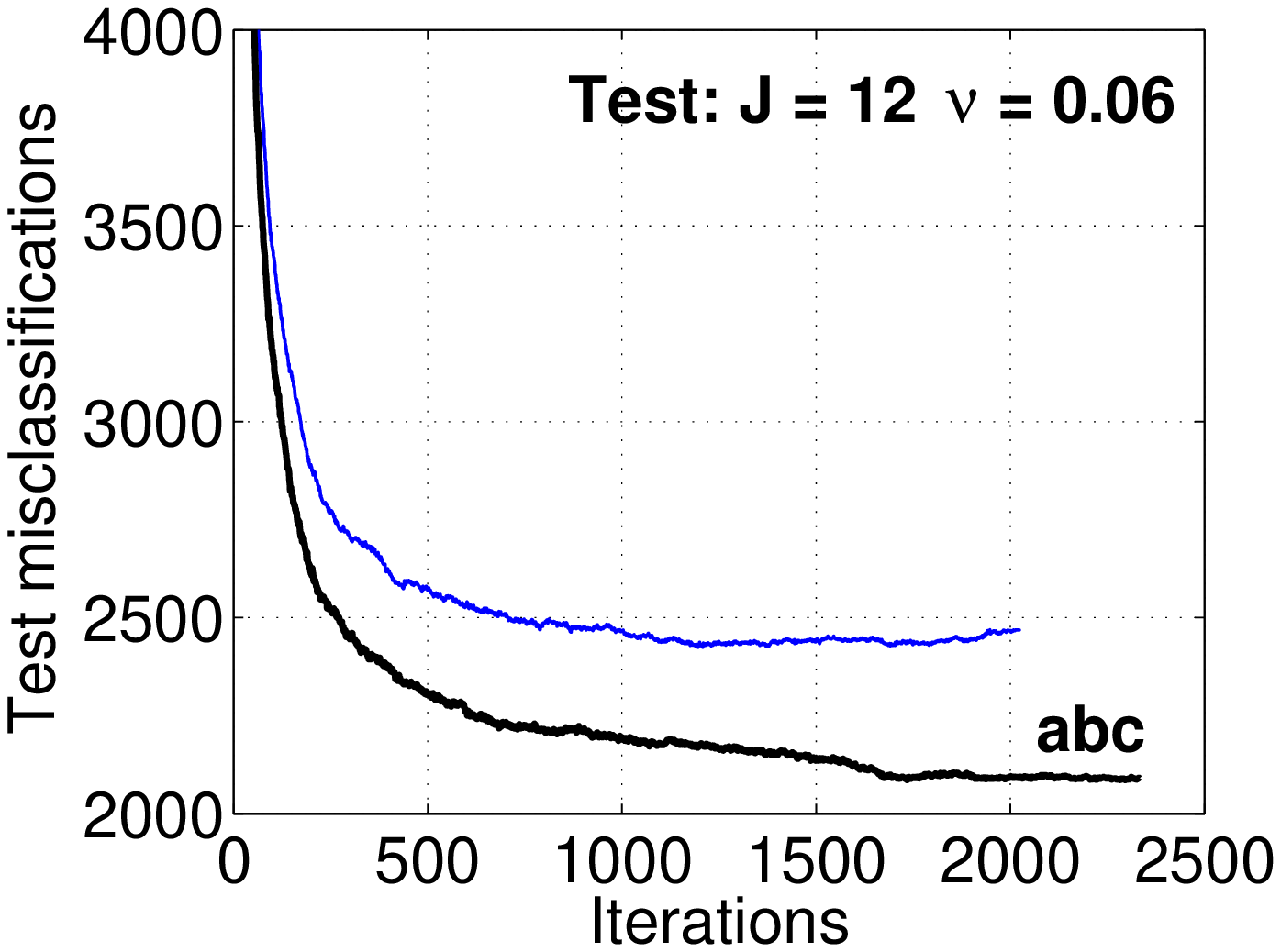}
\includegraphics[width=1.6in]{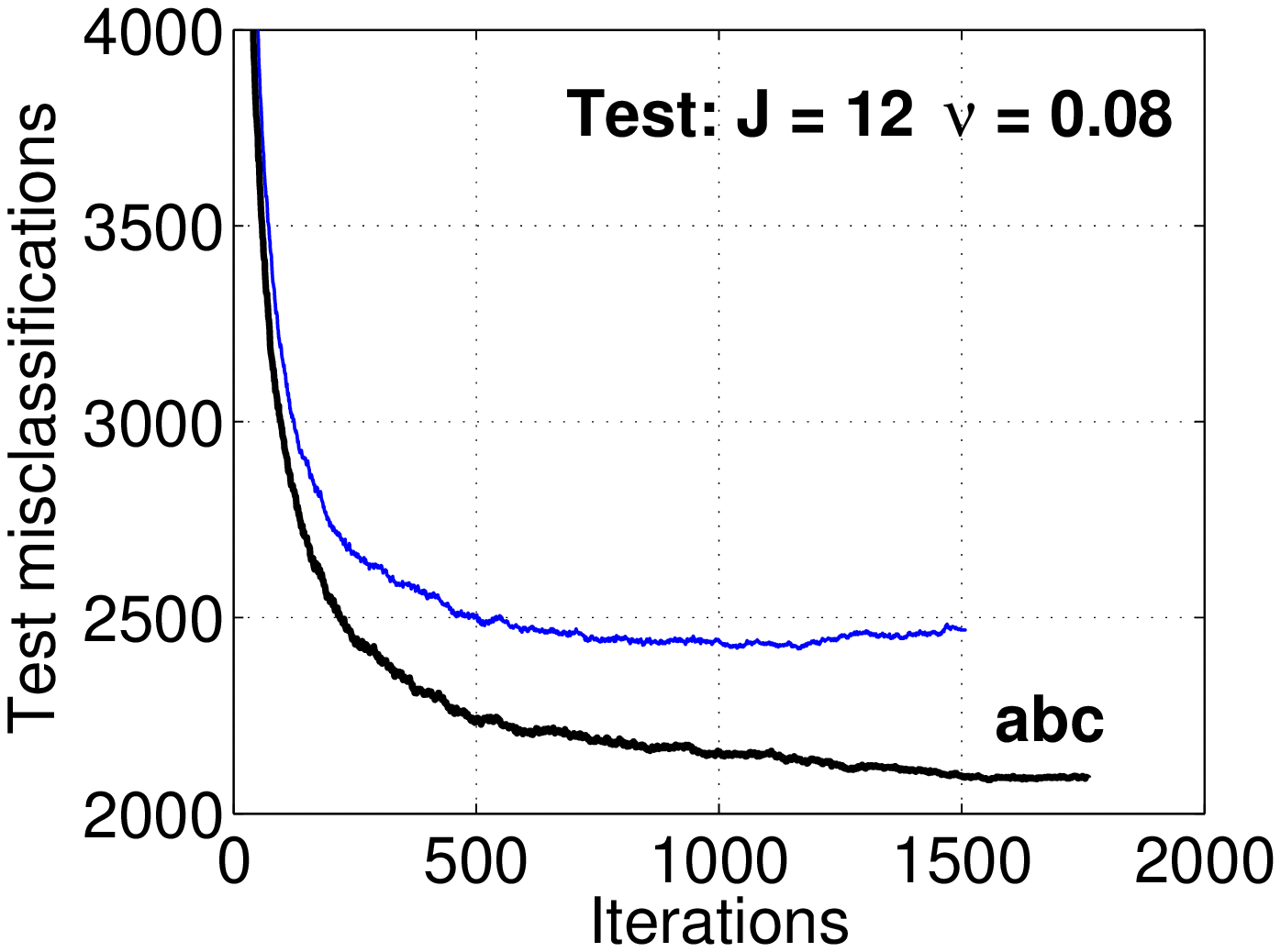}
\includegraphics[width=1.6in]{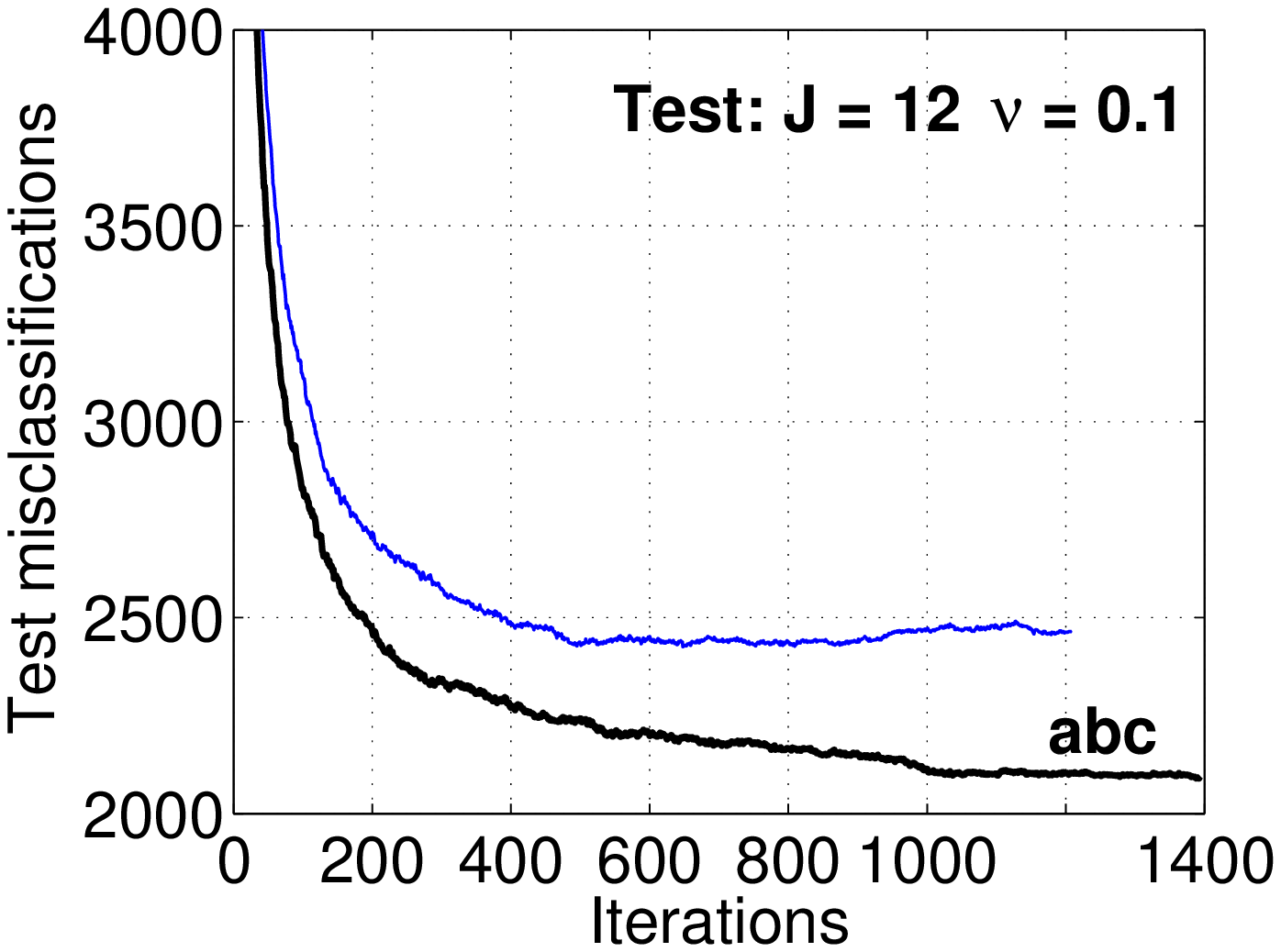}}

\end{center}
\vspace{-0.25in}
\caption{\textbf{\em Mnist10k}. The test mis-classification errors, for {\em logitboost} and {\em abc-logitboost}. $J = 12$ to 20.}\label{fig_Mnist10kTest}
\end{figure}

\begin{figure}[h]
\begin{center}
\mbox{\includegraphics[width=1.6in]{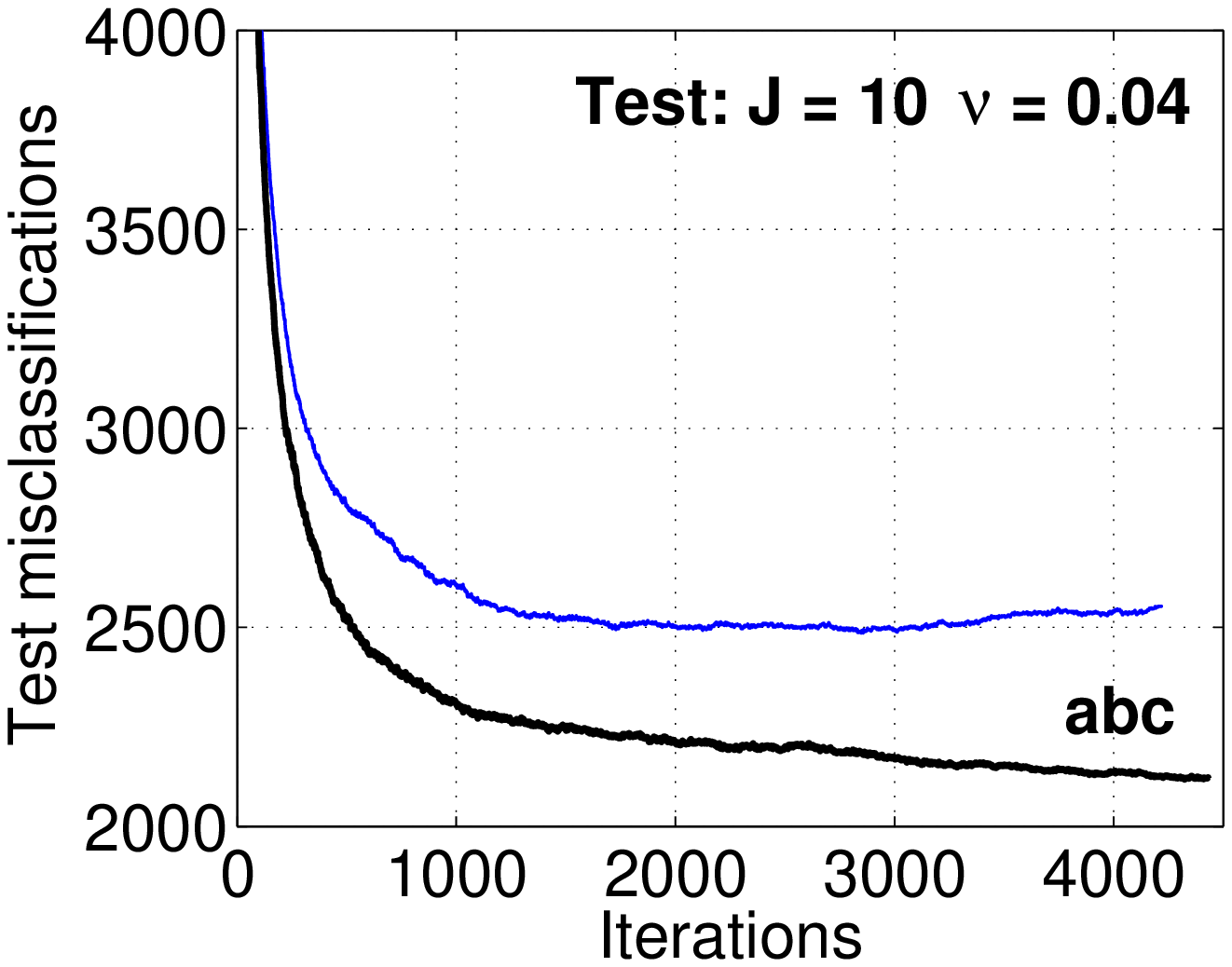}
\includegraphics[width=1.6in]{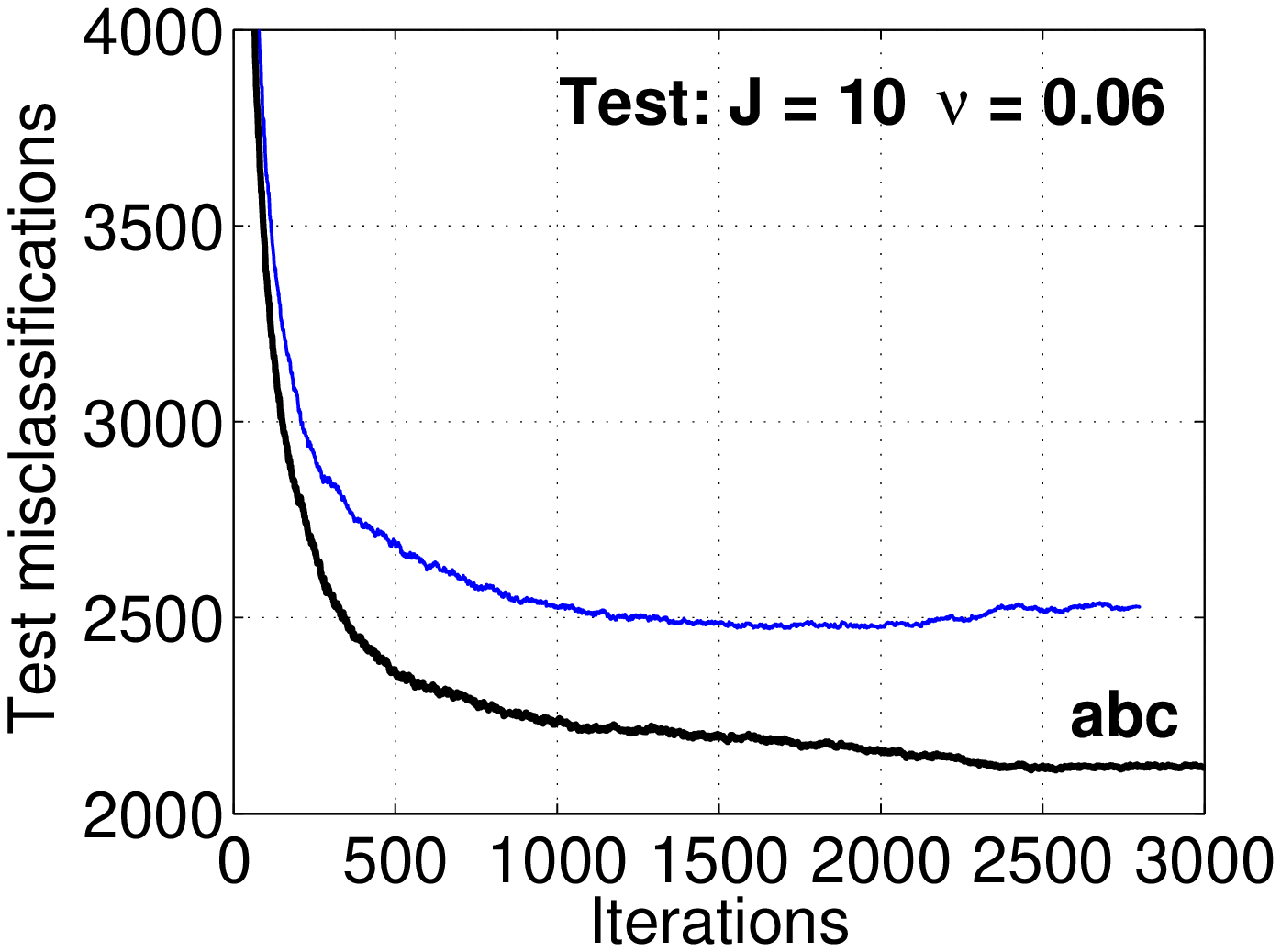}
\includegraphics[width=1.6in]{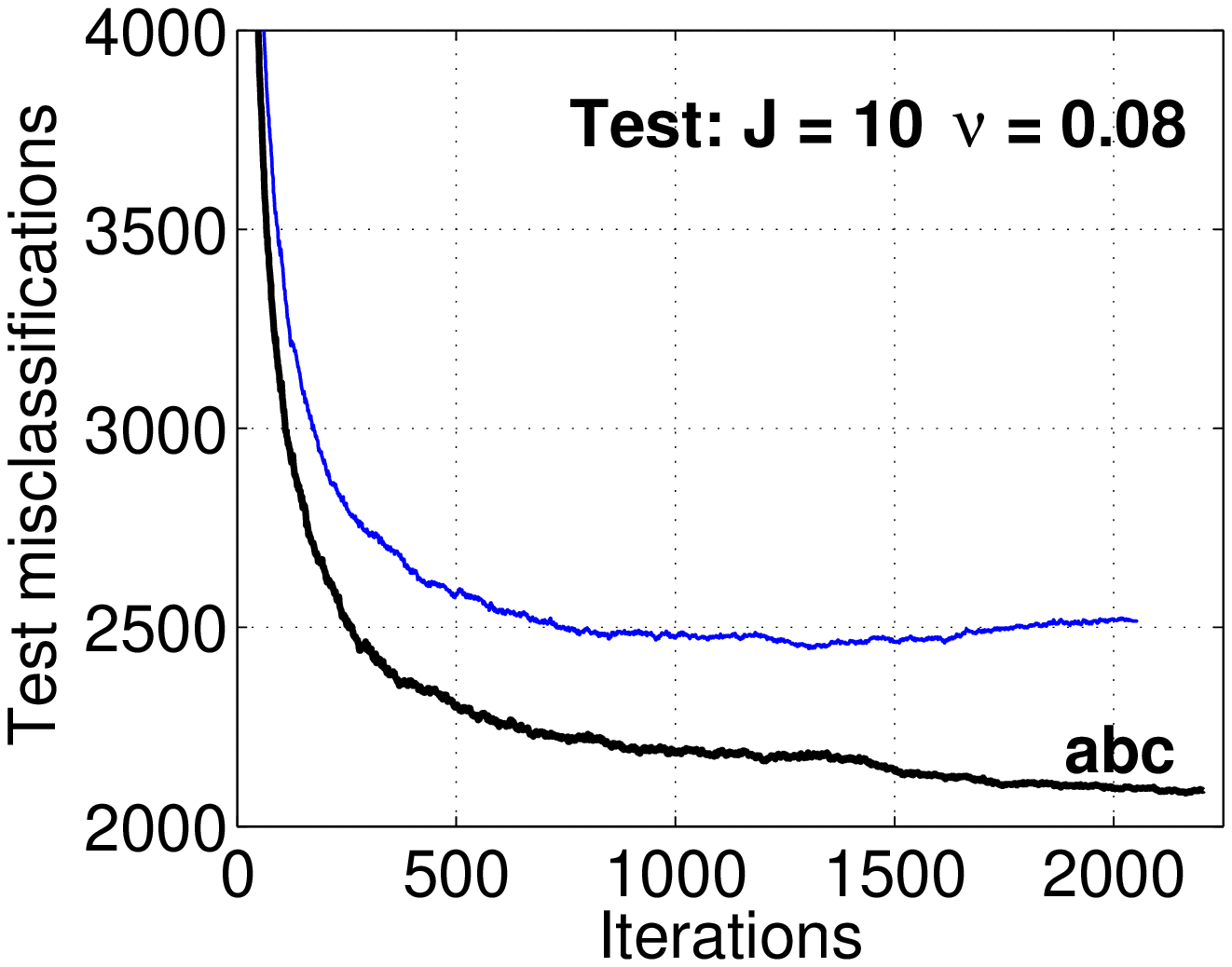}
\includegraphics[width=1.6in]{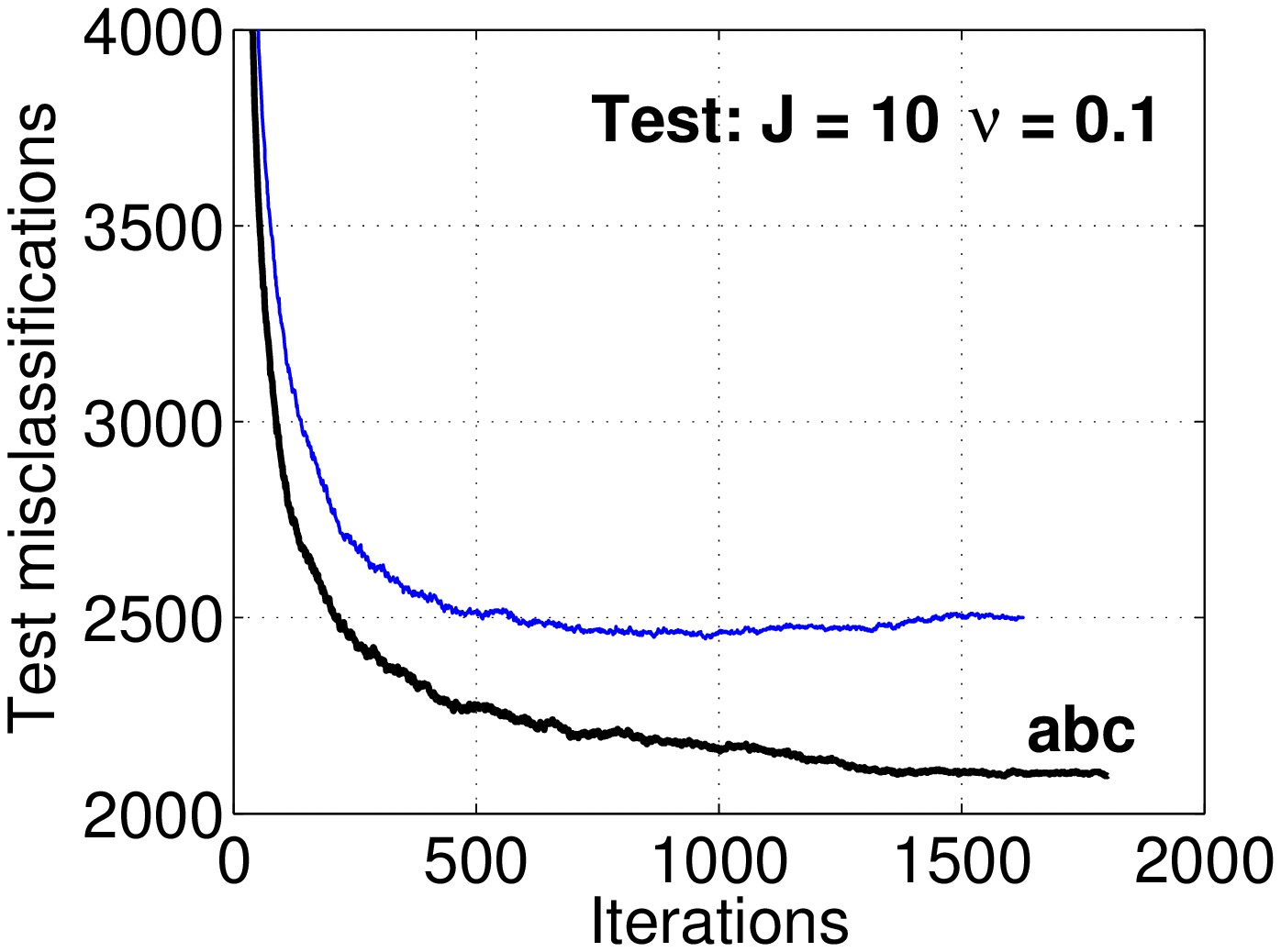}}

\mbox{\includegraphics[width=1.6in]{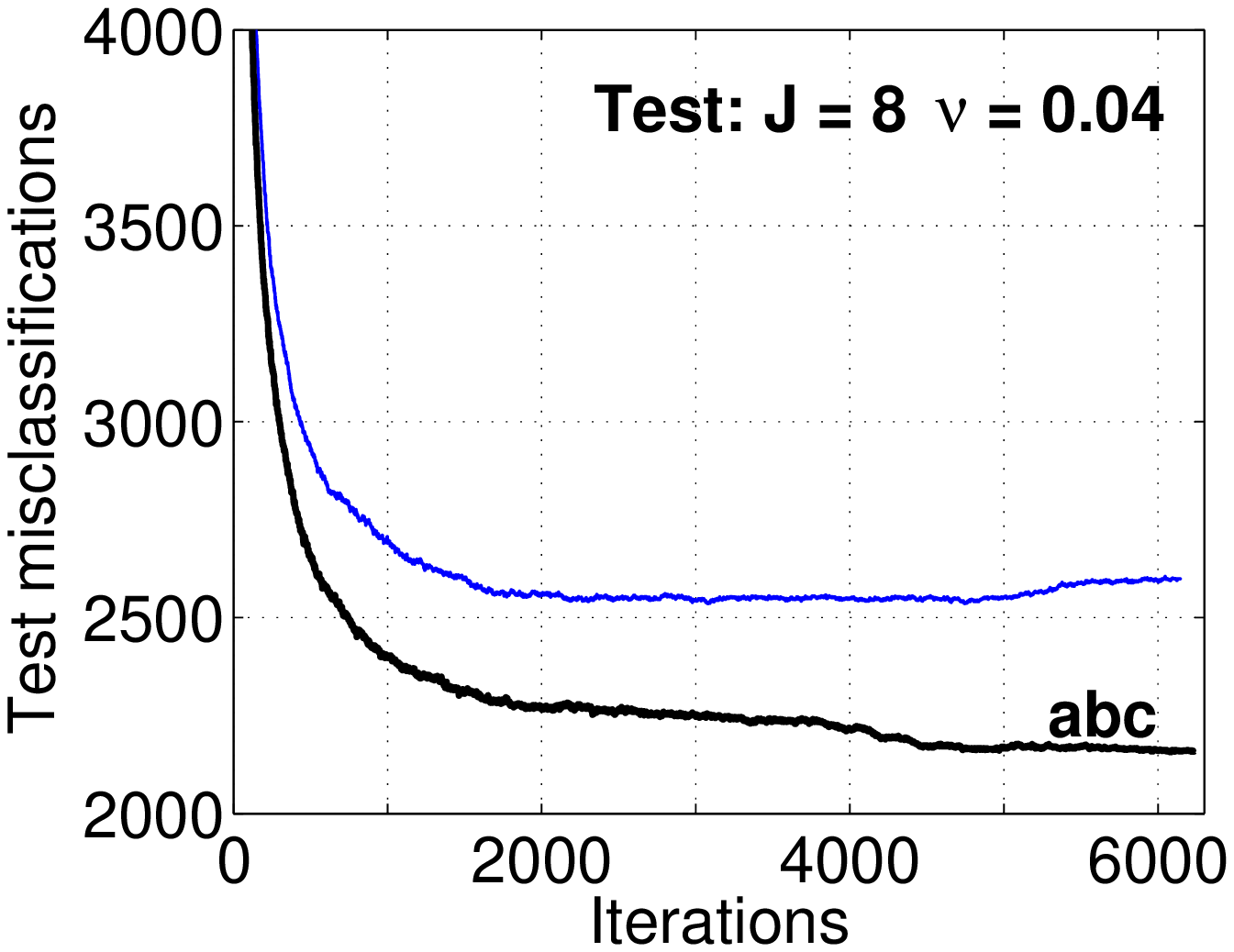}
\includegraphics[width=1.6in]{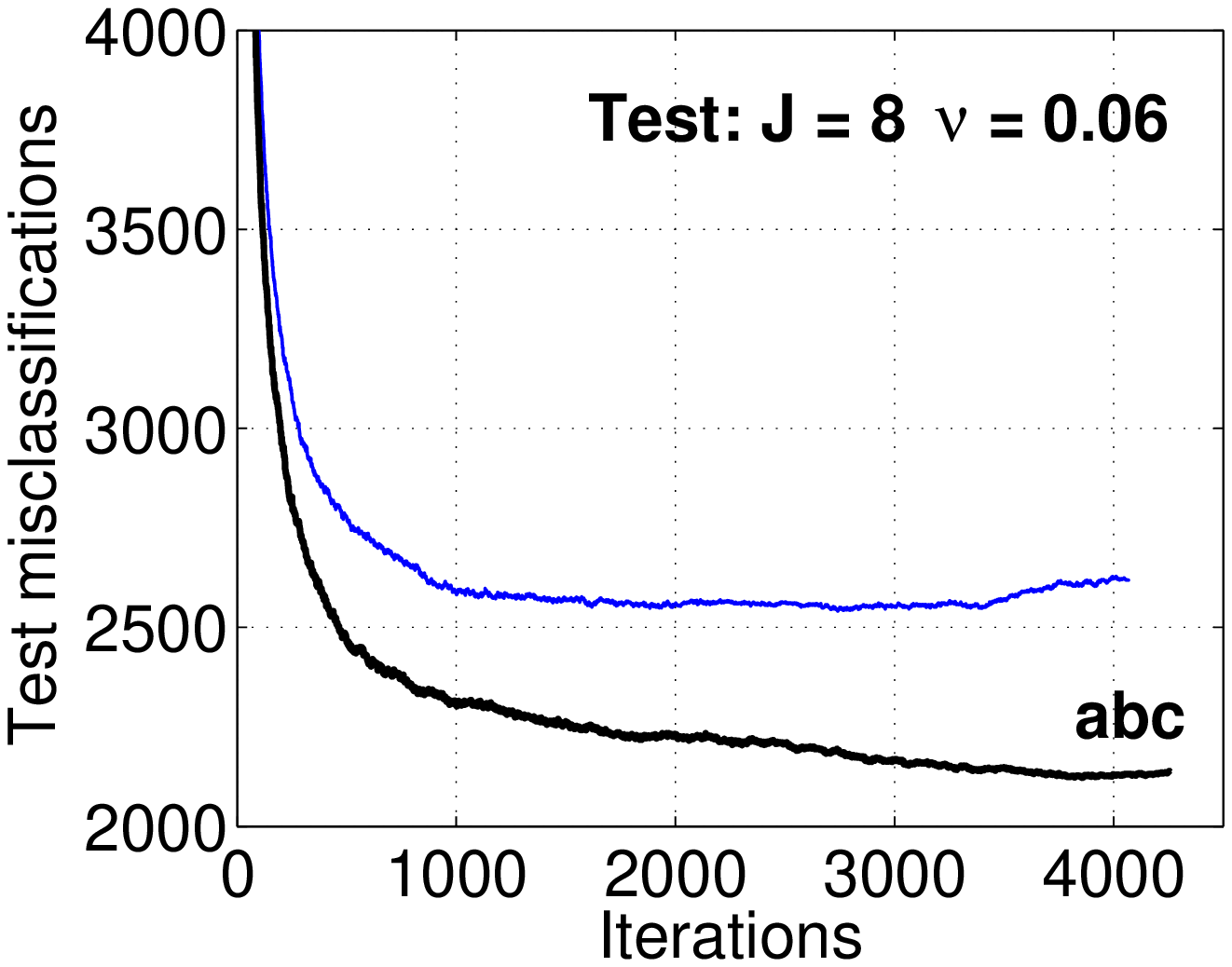}
\includegraphics[width=1.6in]{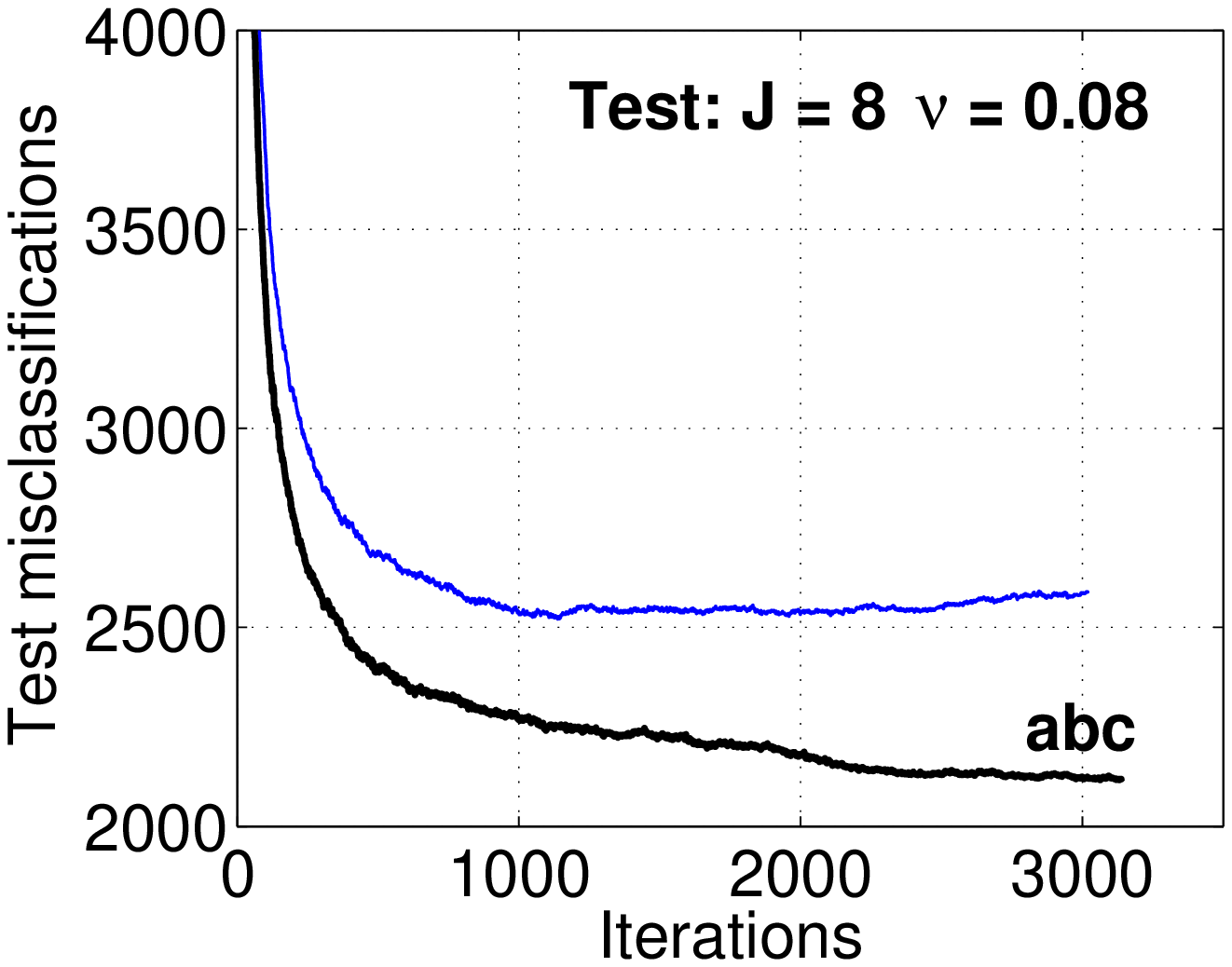}
\includegraphics[width=1.6in]{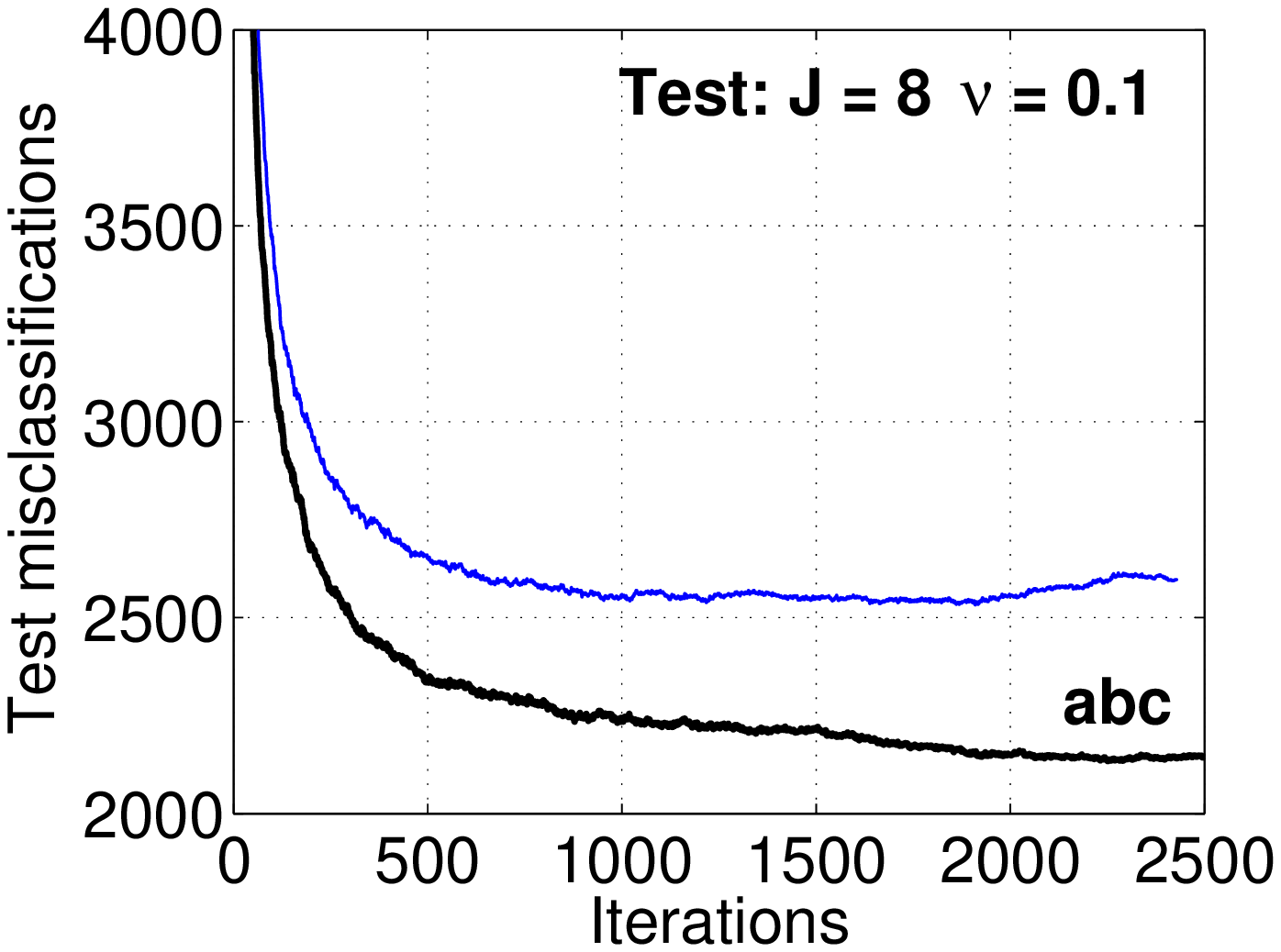}}

\mbox{\includegraphics[width=1.6in]{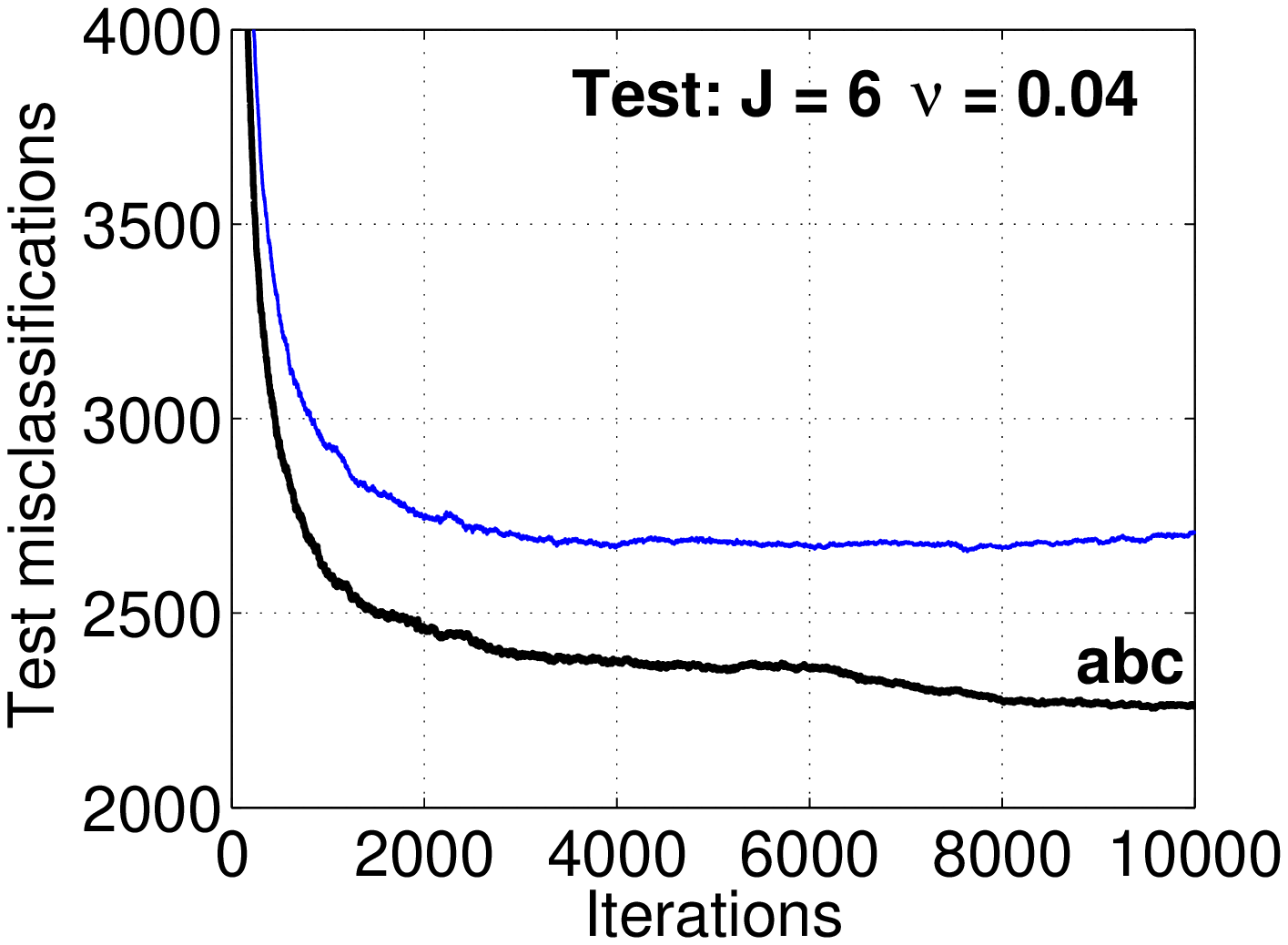}
\includegraphics[width=1.6in]{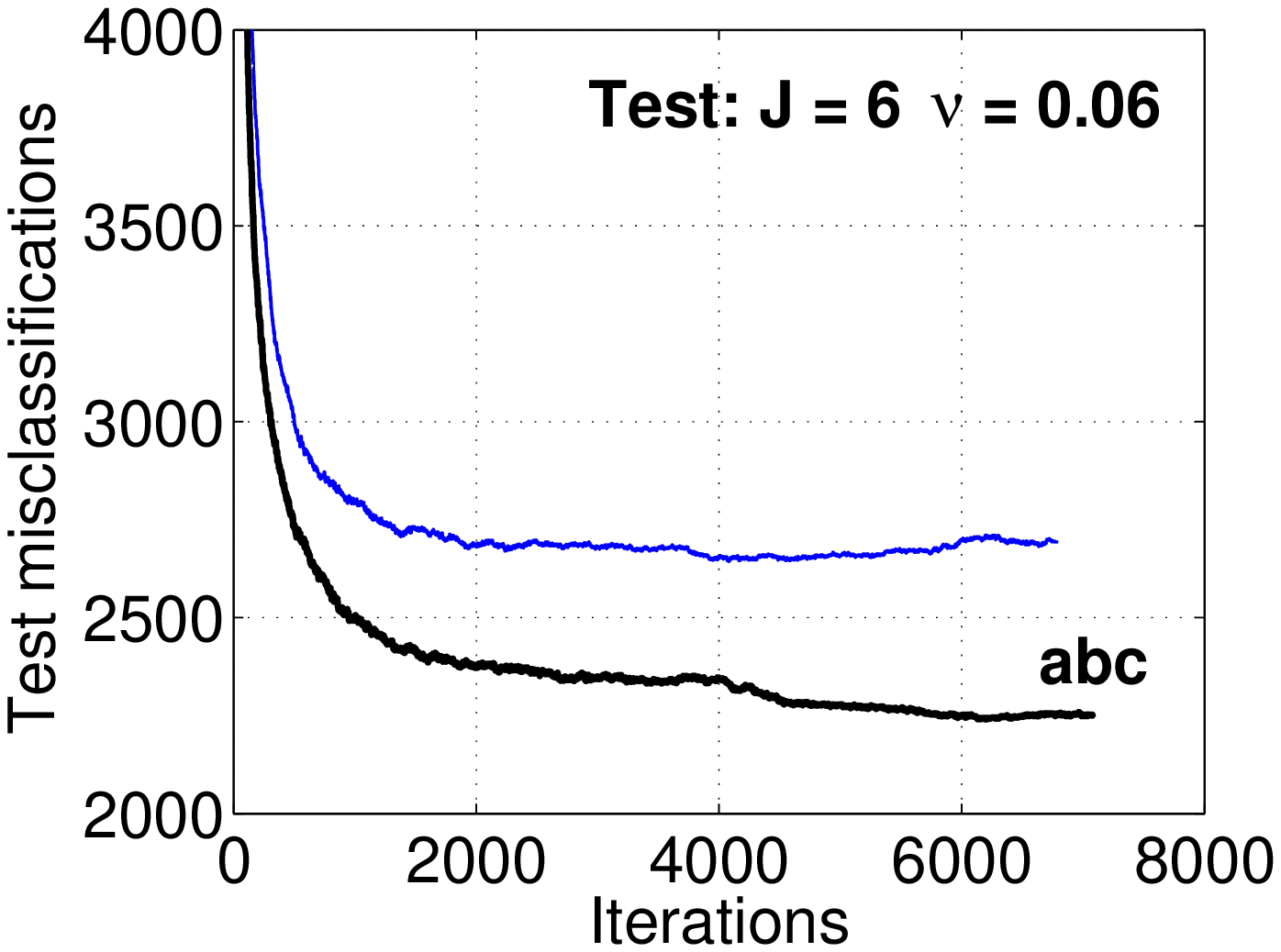}
\includegraphics[width=1.6in]{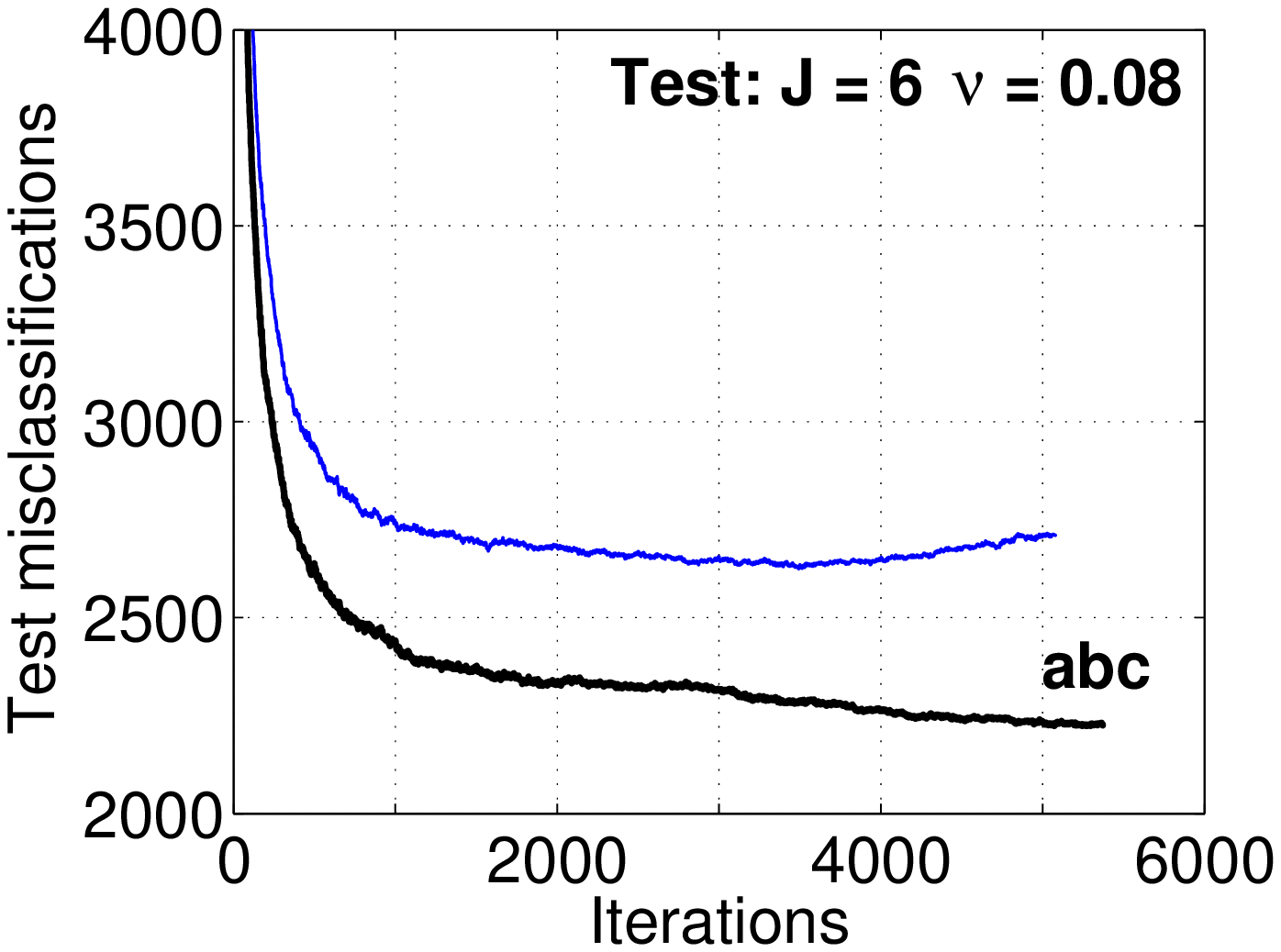}
\includegraphics[width=1.6in]{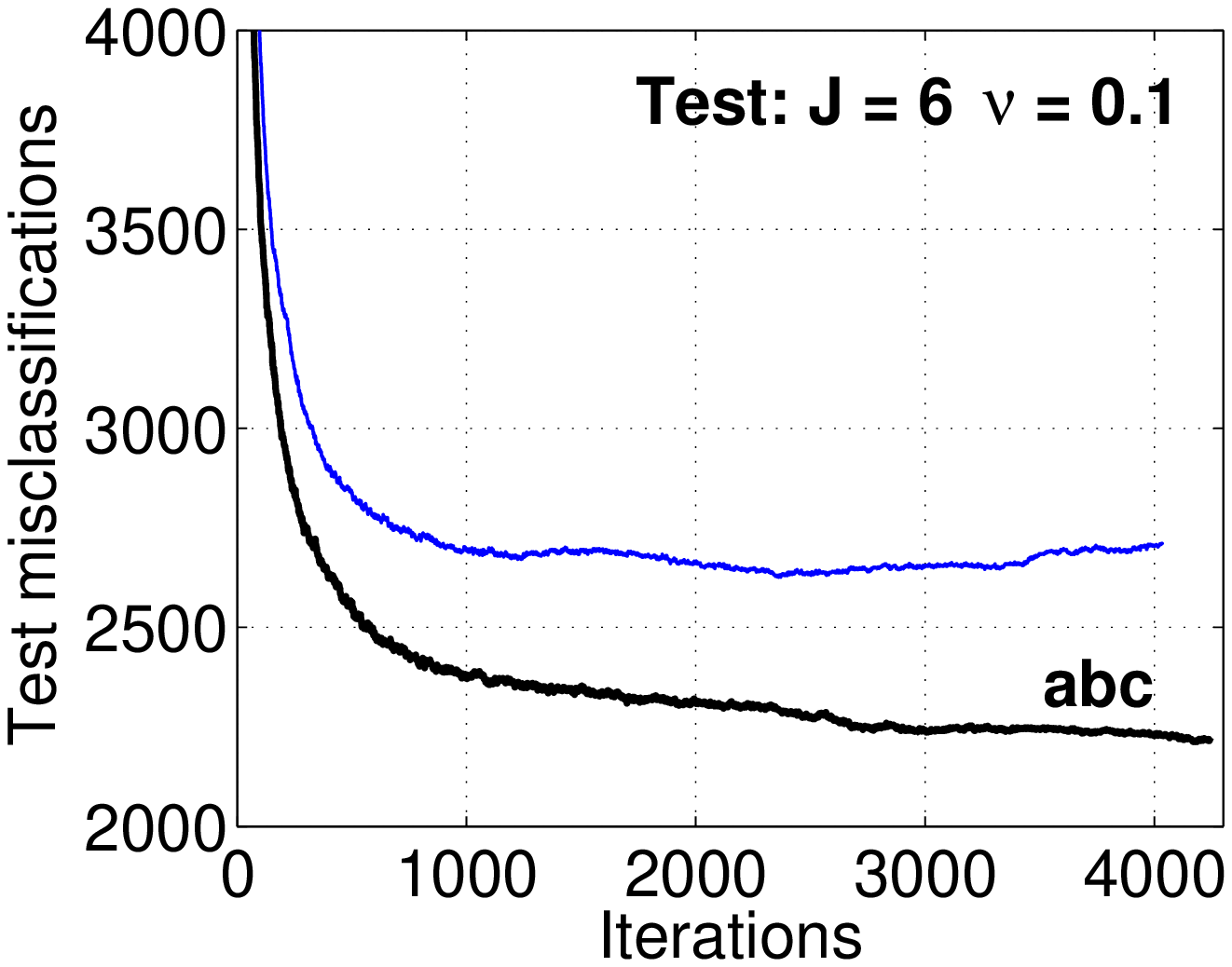}}

\mbox{\includegraphics[width=1.6in]{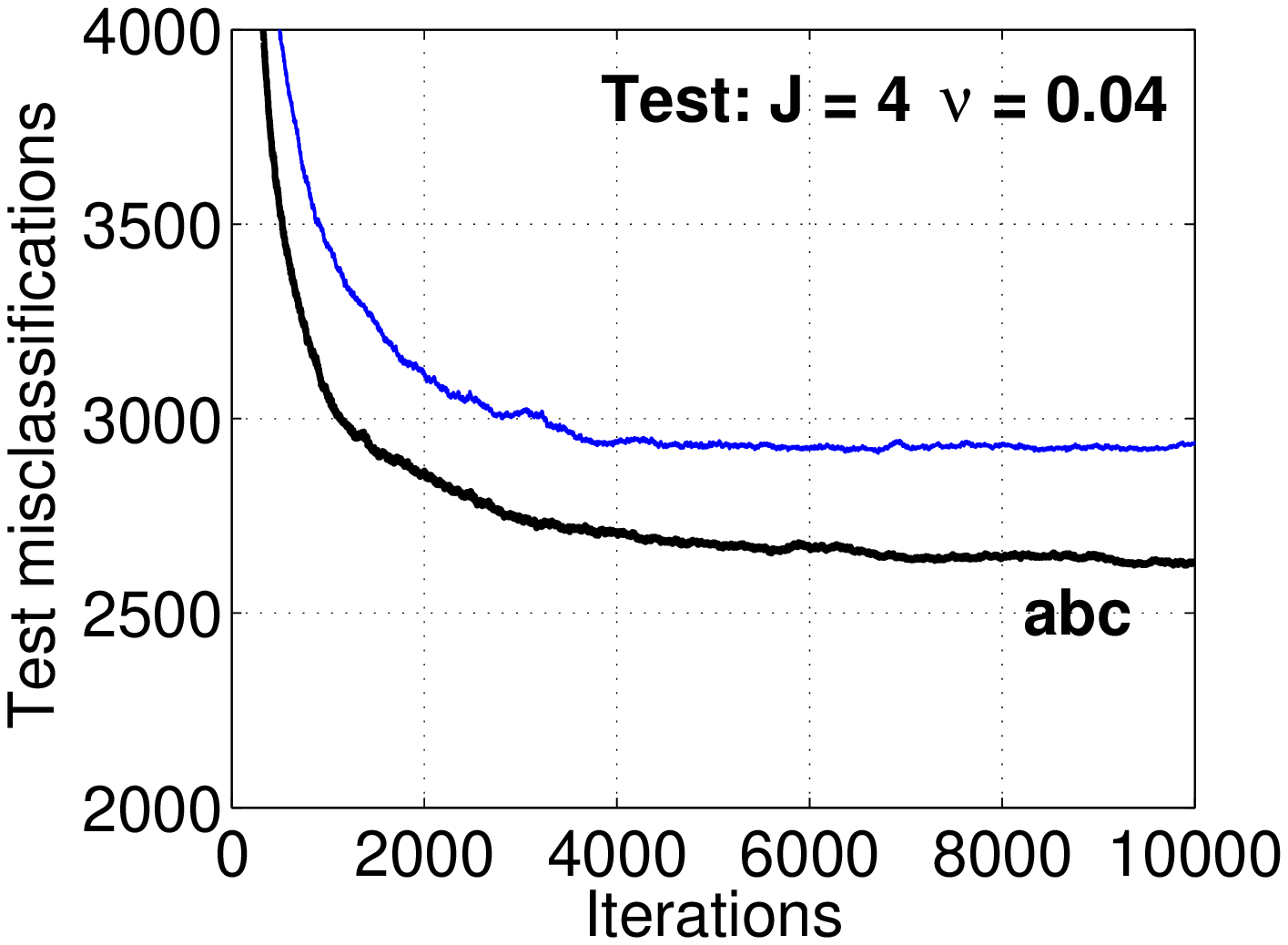}
\includegraphics[width=1.6in]{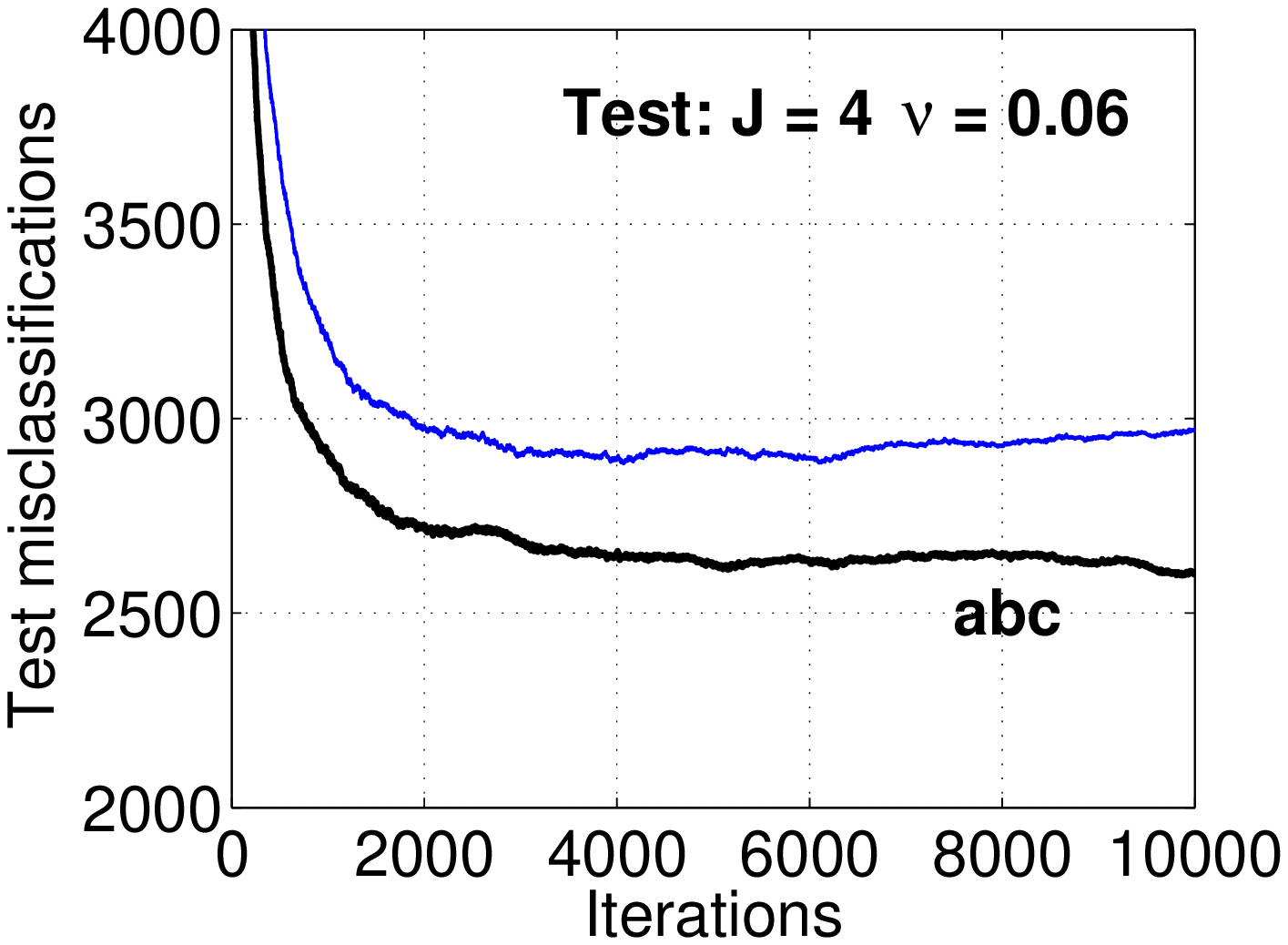}
\includegraphics[width=1.6in]{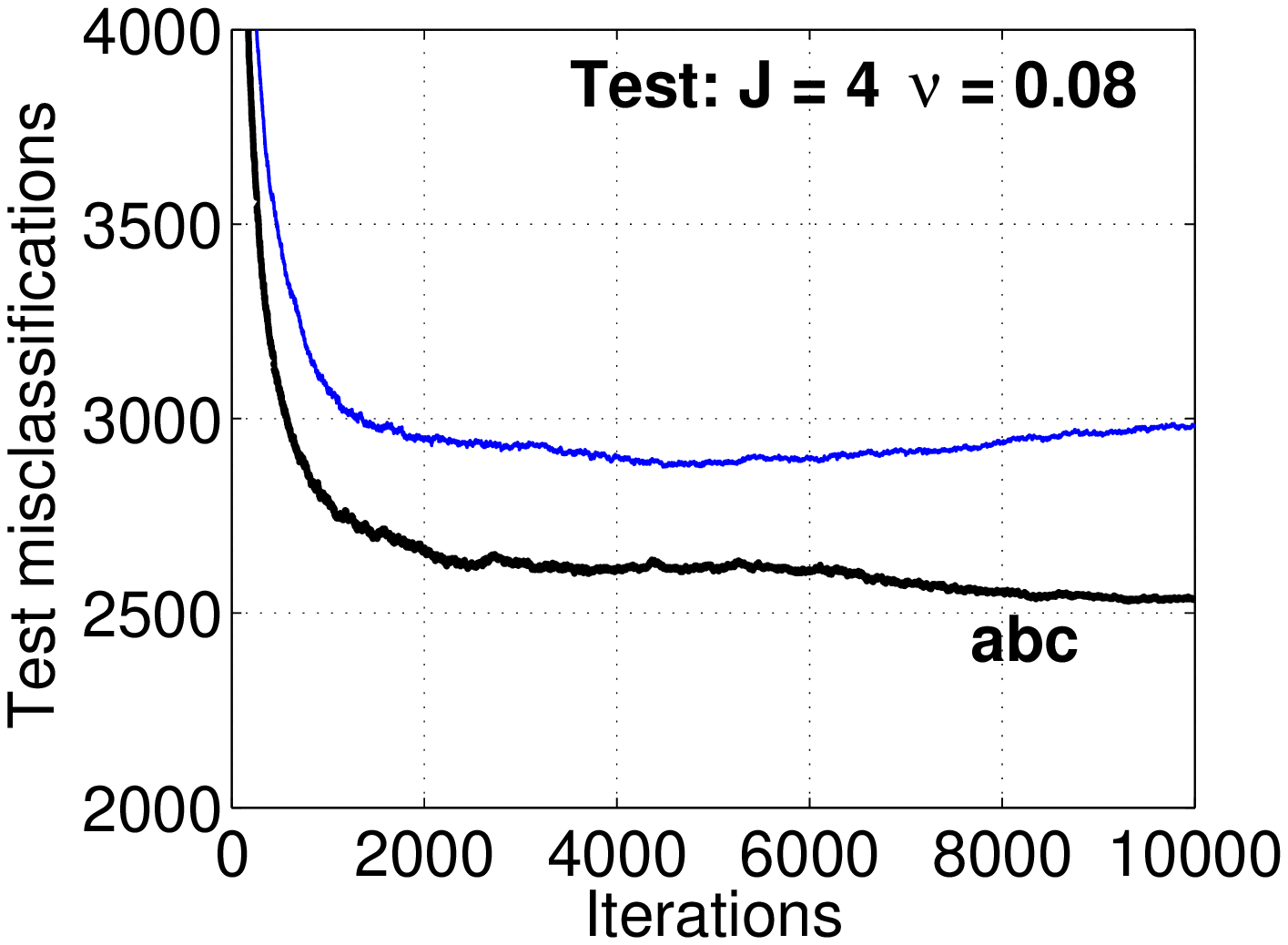}
\includegraphics[width=1.6in]{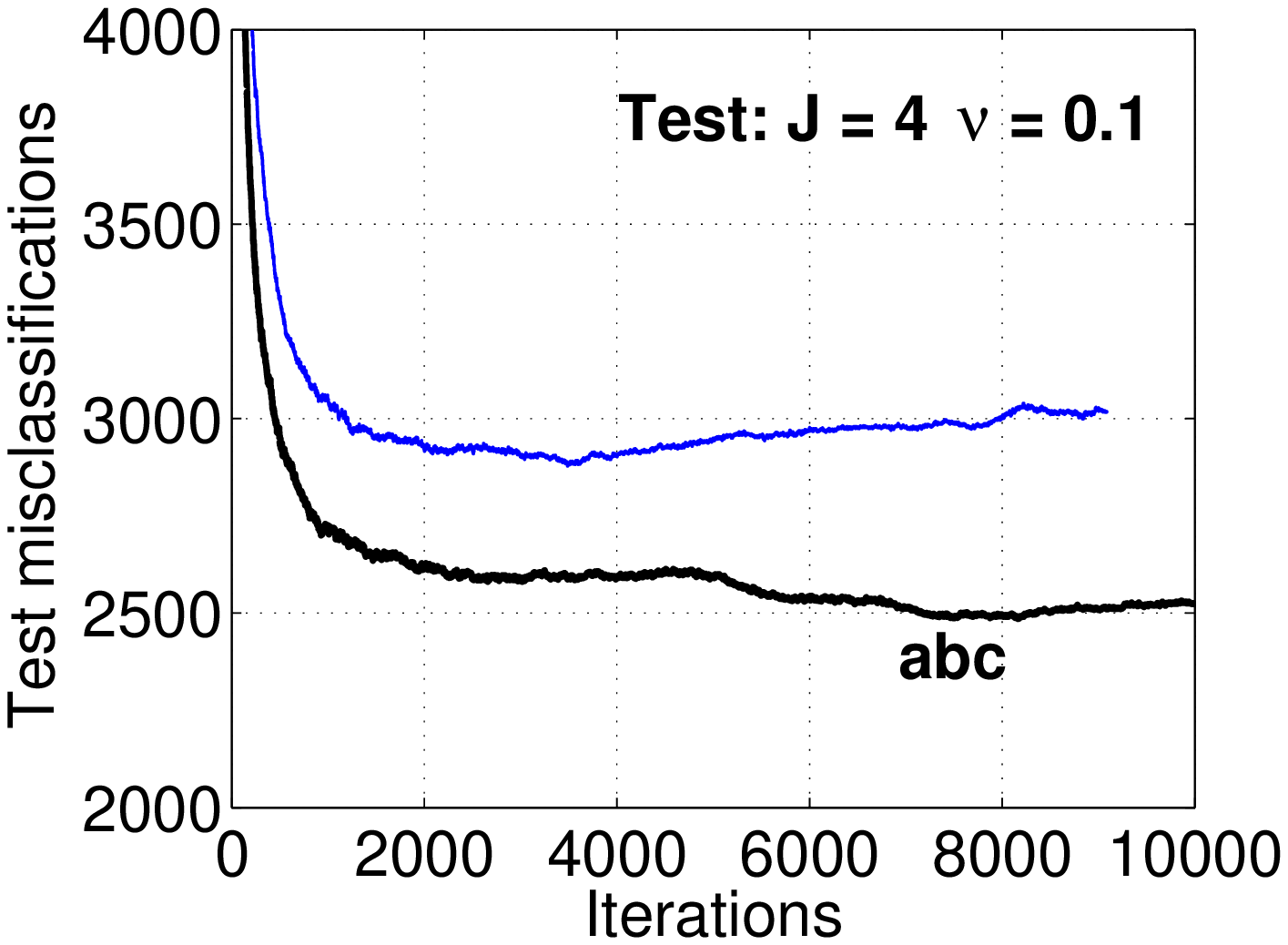}}

\end{center}
\vspace{-0.25in}
\caption{\textbf{\em Mnist10k}. The test mis-classification errors, for {\em logitboost} and {\em abc-logitboost}. $J = 4$ to 10. }\label{fig_Mnist10kTest2}
\end{figure}

Figures \ref{fig_Mnist10kTest} and \ref{fig_Mnist10kTest2} present the mis-classification errors for every $\nu$, $J$, and $M$:
\begin{itemize}
\item Essentially no ovefitting is observed, especially for {\em abc-logitboost}. This is why we simply report the smallest test error in Table \ref{tab_Mnist10k}.
\item The performance is not  sensitive to $\nu$.
\item The performance is not very sensitive to $J$, for $J = 8$ to 20.
\end{itemize}

Interestingly, {\em abc-logitboost} sometimes needed more iterations to reach machine zero than {\em logitboost}. This can be explained in part by the fact that the ``$\nu$'' in {\em logitboost} is not precisely the same  ``$\nu$'' in {\em abc-logitboost}\cite{Proc:ABC_ICML09}. This is also why we would like to experiment with a range of $\nu$ values.

Table \ref{tab_Mnist10k} summarizes the smallest test mis-classification errors along with the relative improvements (denoted by $R_{err}$) of {\em abc-logitboost} over {\em logitboost}. For most $J$ and $\nu$, {\em abc-logitboost} exhibits about $R_{err} = 12\sim 15 (\%)$ smaller test mis-classification errors than {\em logitboost}. The $P$-values range from $1.9\times10^{-10}$ to $3.9\times10^{-5}$, although they are not reported in Table \ref{tab_Mnist10k}.

\begin{table}[h]
\caption{\textbf{\em Mnist10k}. The test mis-classification errors of  {\em logitboost} and \textbf{\em abc-logitboost}, along with the relative improvement $R_{err}$ ($\%$).  For each $J$ and $\nu$, we report the smallest values in Figures \ref{fig_Mnist10kTest} and \ref{fig_Mnist10kTest2}. Each cell contains three numbers, which are {\em logitboost error}, \textbf{\em abc-logitboost error}, and relative improvement $R_{err}$ ($\%$).}
\begin{center}
{\small
{\begin{tabular}{l l l l l }
\hline \hline
  &$\nu = 0.04$ &$\nu=0.06$ &$\nu=0.08$ &$\nu=0.1$ \\
\hline
$J=4$ &2911\ \textbf{2623}\ \ 9.9    &2884\  \textbf{2597}\ 10.0 &2876\ \textbf{2530}\ 12.0   &2878\  \textbf{2485}\ 13.7\\
$J=6$ &2658\ \textbf{2255}\ 15.2   &2644\  \textbf{2240}\ 15.3   &2625\ \textbf{2224}\ 15.3   &2626\  \textbf{2212}\ 15.8\\
$J=8$ &2536\ \textbf{2157}\ 14.9   &2541\  \textbf{2122}\ 16.5   &2521\ \textbf{2117}\ 16.0   &2533\  \textbf{2134}\ 15.8\\
$J=10$ &2486\ \textbf{2118}\ 14.8   &2472\  \textbf{2111}\ 14.6  &2447\ \textbf{2083}\ 14.9   &2446\  \textbf{2095}\ 14.4\\
$J=12$ &2435\ \textbf{2082}\ 14.5    &2424\  \textbf{2086}\ 13.9 &2420\  \textbf{2086}\ 13.8  &2426\  \textbf{2090}\ 13.9\\
$J=14$ &2399\ \textbf{2083}\ 13.2  &2407\   \textbf{2081}\ 13.5  &2402\ \textbf{2056}\  14.4  &2400\   \textbf{2048}\ 14.7\\
$J=16$ &2421\ \textbf{2098}\ 13.3   &2405\  \textbf{2114}\ 12.1  &2382\ \textbf{2083}\  12.6  &2364\  \textbf{2079}\ 12.1\\
$J=18$ &2397\ \textbf{2086}\ 13.0  &2397\  \textbf{2079}\  13.3  &2386\  \textbf{2080}\ 12.8  &2357\  \textbf{2085}\ 11.5\\
$J=20$ &2384\ \textbf{2124}\ 10.9  &2409\  \textbf{2109}\  14.5  &2404\  \textbf{2095}\ 12.9  &2372\ \textbf{2101}\ 11.4
\\\hline\hline
\end{tabular}}}
\end{center}
\label{tab_Mnist10k}
\end{table}

\vspace{1in}

The original {\em abc-boost} paper\cite{Proc:ABC_ICML09} did not include experiments on {\em Mnist10k}. Thus, in this study,
Table \ref{tab_Mnist10k_mart} summarizes the smallest test mis-classification errors for {\em mart} and {\em abc-mart}. Again, we can see very consistent and considerable improvement of {\em abc-mart} over {\em mart}. Also, comparing Tables \ref{tab_Mnist10k} and \ref{tab_Mnist10k_mart}, we can see that {\em abc-logitboost} also significantly improves {\em abc-mart}.

\begin{table}[h]
\caption{\textbf{\em Mnist10k}. The test mis-classification errors of  {\em mart} and \textbf{\em abc-mart}, along with the relative improvement $R_{err}$ ($\%$).  For each $J$ and $\nu$, we report the smallest values in Figures \ref{fig_Mnist10kTest} and \ref{fig_Mnist10kTest2}. Each cell contains three numbers, which are {\em mart error}, \textbf{\em abc-mart error}, and relative improvement $R_{err}$ ($\%$).}%\vspace{-0.2in}
\begin{center}
{\small
{\begin{tabular}{l l l l l }
\hline \hline
  &$\nu = 0.04$ &$\nu=0.06$ &$\nu=0.08$ &$\nu=0.1$ \\
\hline
$J=4$ &3346\ \textbf{3054}\ \ 8.7    &3308\  \textbf{3009}\ \ 9.0 &3302\ \textbf{2855}\ 13.5   &3287\  \textbf{2792}\ 15.1\\
$J=6$ &3176\ \textbf{2752}\ 13.4   &3074\  \textbf{2624}\ 14.6   &3071\ \textbf{2649}\ 13.7   &3089\  \textbf{2572}\ 16.7\\
$J=8$ &3040\ \textbf{2557}\ 15.9   &3012\  \textbf{2552}\ 15.2   &3000\ \textbf{2529}\ 15.7   &2993\  \textbf{2566}\ 14.3\\
$J=10$ &2979\ \textbf{2537}\ 14.8   &2941\  \textbf{2515}\ 14.5  &2957\ \textbf{2509}\ 15.2   &2947\  \textbf{2493}\ 15.4\\
$J=12$ &2912\ \textbf{2498}\ 14.2    &2897\  \textbf{2453}\ 15.3 &2906\  \textbf{2475}\ 14.8  &2887\  \textbf{2469}\ 14.5\\
$J=14$ &2907\ \textbf{2473}\ 14.9  &2886\   \textbf{2466}\ 14.6  &2874\ \textbf{2463}\  14.3  &2864\   \textbf{2435}\ 15.0\\
$J=16$ &2885\ \textbf{2466}\ 14.5   &2879\  \textbf{2441}\ 15.2  &2868\ \textbf{2459}\  14.2  &2854\  \textbf{2451}\ 14.1\\
$J=18$ &2852\ \textbf{2467}\ 13.5  &2860\  \textbf{2447}\  14.4  &2865\  \textbf{2436}\ 15.0  &2852\  \textbf{2448}\ 14.2\\
$J=20$ &2831\ \textbf{2438}\ 13.9  &2833\  \textbf{2440}\  13.9  &2832\  \textbf{2425}\ 14.4  &2813\ \textbf{2434}\ 13.5
\\\hline\hline
\end{tabular}}}
\end{center}
\label{tab_Mnist10k_mart}
\end{table}

\clearpage

\subsection{Summary of Test Mis-Classification Errors}

Table \ref{tab_summary} summarizes the test errors, which are the overall best (smallest) test mis-classification errors. In the table, $R_{err}$ ($\%$) is the relative improvement of test performance. The $P$-values tested the statistical significance if {\em abc-logitboost} achieved smaller \textbf{error rates} than {\em logitboost}.

To compare {\em abc-logitboost} with {\em abc-mart}, Table \ref{tab_summary} also includes the test errors for {\em abc-mart} and the $P$-values (i.e., $P$-value (2)) for testing the statistical significance if {\em abc-logitboost} achieved smaller \textbf{error rates} than {\em abc-mart}. The comparisons indicate that there is a clear performance gap between {\em abc-logitboost} and {\em abc-mart}, especially on the large datasets.

\begin{table}[h]
\caption{Summary of test mis-classification errors.   }
\begin{center}{\small
\begin{tabular}{l r r r r r r r}
\hline \hline
Dataset &\textbf{logit} & \textbf{abc-logit} & $R_{err}$ (\%)  &$P$-value &\hspace{0.2in}  & \textbf{abc-mart} & $P$-vlaue (2) \\
\hline
Covertype &10759 & 9693 & 9.9 & $1.6\times10^{-14}$& & 10375 &$4.8\times10^{-7}$ \\
Mnist10k &2357 &2048& 13.1 & $1.0\times 10^{-6}$& &2425 &$4.6\times10^{-9}$\\
Letter2k &2257   & 1984   &12.1 &$4.0\times 10^{-6}$&& 2180 &$6.2\times10^{-4}$ \\
Letter4k &1220   & 1031   &15.5 &$1.8\times 10^{-5}$& & 1126 &0.017\\
Letter &107   & 89   &16.8 &$9.7\times 10^{-3}$&& 99 &0.23\\
Pendigits &109 &90   &17.4 &$8.6\times10^{-3}$&&100 &0.23 \\
Zipcode  &103  &92   &10.7 &0.21& &100 &0.28\\
Optdigits &49  &38   &22.5 &0.11& &43  &0.29\\
Isolet &62   & 55   &11.3 &0.25 & & 64 &0.20 \\
\hline\hline
\end{tabular}
}
\end{center}
\label{tab_summary}
\end{table}

\subsection{Experiments on the {\em Covertype} Dataset }\label{sec_Covertype}

Table \ref{tab_Covertype} summarizes the smallest test mis-classification errors of {\em logitboost} and {\em abc-logitboost}, along with the relative improvements ($R_{err}$).  Since this is a fairly large dataset, we only experimented with $\nu=0.1$ and $J=10$ and $20$.

\begin{table}[h]
\caption{\textbf{\em Covertype}. We report the test mis-classification errors of {\em logitboost} and {\em abc-logitboost}, together with the relative improvements ($R_{err}$, $\%$) in parentheses. }
\begin{center}
{\small
%\subtable[ABC-MART]
{\begin{tabular}{l l l l l }
\hline \hline
$\nu$ &$M$ &$J$  &logit  &abc-logit \\
\hline
0.1 &1000 &10     &29865        &23774 (20.4)\\
0.1 &1000 &20       &19443       &14443 (25.7)\\\hline
0.1 &2000 &10     &21620        &16991 (21.4)\\
0.1 &2000 &20       &13914     &11336 (18.5)\\\hline
0.1 &3000 &10 &17805      &14295 (19.7)\\
0.1 &3000 &20 &12076         & 10399 (13.9)\\\hline
0.1 &5000 &10   &14698      &12185 (17.1)   \\
0.1 &5000&20       &10759      &\ 9693 \ (9.9) \\\hline\hline
\end{tabular}}
}
\end{center}
\label{tab_Covertype}%\vspace{-0.25in}
\end{table}

The results on {\em Covertype} are reported differently from other datasets. {\em Covertype} is fairly large. Building a very large model (e.g., $M=5000$ boosting steps) would be expensive. Testing a very large model at run-time can be  costly or infeasible for certain applications (e.g., search engines). Therefore, it is often important to examine the performance of the algorithm at much earlier boosting iterations. Table \ref{tab_Covertype} shows that {\em abc-logitboost} may improve {\em logitboost} as much as  $R_{err}\approx20\%$, as opposed to the reported $R_{err}=9.9\%$ in Table \ref{tab_summary}.

\clearpage

\subsection{Experiments on the {\em  Letter2k} Dataset}

\begin{table}[h]
\caption{\textbf{\em Letter2k}. The test mis-classification errors of  {\em logitboost} and \textbf{\em abc-logitboost}, along with the relative improvement $R_{err}$ ($\%$).  Each cell contains three numbers, which are {\em logitboost error}, \textbf{\em abc-logitboost error}, and relative improvement $R_{err}$ ($\%$).}%\vspace{-0.2in}
\begin{center}
{\small
{\begin{tabular}{l l l l l }
\hline \hline
  &$\nu = 0.04$ &$\nu=0.06$ &$\nu=0.08$ &$\nu=0.1$ \\
\hline
$J=4$ &2576\ \textbf{2317}\ 10.1    &2535\  \textbf{2294}\ \ 9.5 &2545\ \textbf{2252}\ 11.5   &2523\  \textbf{2224}\ 11.9\\
$J=6$ &2389\ \textbf{2133}\  10.7    &2391\  \textbf{2111}\ 11.7 &2376\ \textbf{2070}\ 12.9   &2370\  \textbf{2064}\ 12.9\\
$J=8$ &2325\ \textbf{2074}\ 10.8   &2299\  \textbf{2046}\ 11.0   &2298\  \textbf{2033}\ 11.5   &2271\   \textbf{2025}\ 10.8\\
$J=10$ &2294\ \textbf{2041}\ 11.0   &2292\  \textbf{1995}\ 13.0  &2279\  \textbf{2018}\ 11.5   &2276\   \textbf{2000}\ 12.1\\
$J=12$ &2314\  \textbf{2010}\ 13.1    &2304\  \textbf{1990}\ 13.6 &2311\  \textbf{2010}\ 13.0  &2268\  \textbf{2018}\ 11.0\\
$J=14$ &2315\   \textbf{2015}\ 13.0  &2300\   \textbf{2003}\ 12.9  &2312\  \textbf{2003}\  13.4  &2277\   \textbf{2024}\ 11.1\\
$J=16$ &2302\  \textbf{2022}\ 12.2   &2394\   \textbf{1996}\ 13.0  &2276\  \textbf{3002}\  12.0  &2257\  \textbf{1997}\ 11.5\\
$J=18$ &2295\  \textbf{2041}\ 11.1  &2275\  \textbf{2021}\  11.2  &2301\   \textbf{1984}\ 13.8  &2281\  \textbf{2020}\ 11.4\\
$J=20$ &2280\  \textbf{2047}\ 10.2  &2267\  \textbf{2020}\  10.9  &2294\  \textbf{2020}\ 11.9  &2306\  \textbf{2031}\ 11.9
\\\hline\hline
\end{tabular}}}
\end{center}
\label{tab_Letter2k}\vspace{-0.in}
\end{table}

\vspace{0.5in}

\subsection{Experiments on the {\em  Letter4k} Dataset}
\vspace{-0in}

\begin{table}[h]
\caption{\textbf{\em Letter4k}. The test mis-classification errors of  {\em logitboost} and \textbf{\em abc-logitboost}, along with the relative improvement $R_{err}$ ($\%$). }%\vspace{-0.2in}
\begin{center}
{\small
{\begin{tabular}{l l l l l }
\hline \hline
  &$\nu = 0.04$ &$\nu=0.06$ &$\nu=0.08$ &$\nu=0.1$ \\
\hline
$J=4$ &1460\ \textbf{1295}\ 11.3    &1471\  \textbf{1232}\  16.2 &1452\ \textbf{1199}\ 17.4   &1446\  \textbf{1204}\ 16.7\\
$J=6$ &1390\ \textbf{1135}\  18.3    &1394\  \textbf{1116}\ 20.0 &1382\ \textbf{1088}\ 21.3   &1374\  \textbf{1070}\ 22.1\\
$J=8$ &1336\ \textbf{1078}\ 19.3   &1332\  \textbf{1074}\ 19.4   &1311\  \textbf{1062}\ 19.0   &1297\   \textbf{1042}\ 20.0\\
$J=10$ &1289\ \textbf{1051}\ 18.5   &1285\  \textbf{1065}\ 17.1  &1280\  \textbf{1031}\ 19.5   &1273\   \textbf{1046}\ 17.8\\
$J=12$ &1251\  \textbf{1055}\ 15.7    &1247\  \textbf{1065}\ 14.6 &1261\  \textbf{1044}\ 17.2  &1243\  \textbf{1051}\ 15.4\\
$J=14$ &1247\   \textbf{1060}\ 15.0  &1233\   \textbf{1050}\ 14.8  &1251\  \textbf{1037}\  17.1 &1244\   \textbf{1060}\ 14.8\\
$J=16$ &1244\  \textbf{1070}\ 14.0   &1227\   \textbf{1064}\ 13.3  &1231\  \textbf{1044}\  15.2  &1228\  \textbf{1038}\ 15.5\\
$J=18$ &1243\  \textbf{1057}\ 15.0  &1250\  \textbf{1037}\  17.0  &1234\   \textbf{1049}\ 15.0  &1220\  \textbf{1055}\ 13.5\\
$J=20$ &1226\  \textbf{1078}\ 12.0  &1242\  \textbf{1069}\  13.9  &1242\  \textbf{1054}\ 15.1  &1235\  \textbf{1051}\ 14.9
\\\hline\hline
\end{tabular}}}
\end{center}
\label{tab_Letter4k}\vspace{-0.in}
\end{table}
\newpage

\subsection{Experiments on the {\em  Letter} Dataset}

%\vspace{-0.2in}

\begin{table}[h]
\caption{\textbf{\em Letter}.
 The test mis-classification errors of  {\em logitboost} and \textbf{\em abc-logitboost}, along with the relative improvement $R_{err}$ ($\%$).   }%\vspace{-0.2in}
\begin{center}
{\small
{\begin{tabular}{l l l l l }
\hline \hline
  &$\nu = 0.04$ &$\nu=0.06$ &$\nu=0.08$ &$\nu=0.1$ \\
\hline
$J=4$ &149\ \textbf{125}\ \ 16.1    &151\  \textbf{121}\ 19.9 &148\ \textbf{122}\ 17.6   &149\  \textbf{119}\ 20.1\\
$J=6$ &130\ \textbf{112}\ \ 13.8    &132\  \textbf{107}\ 18.9 &133\ \textbf{101}\ 24.1   &129\  \textbf{102}\ 20.9\\
$J=8$ &129\ \textbf{104}\ 19.4   &125\  \textbf{102}\ 18.4   &131\ \ \textbf{93}\ 29.0   &113\ \  \textbf{95}\ 15.9\\
$J=10$ &114\ \textbf{101}\ 11.4   &115\  \textbf{100}\ 13.0  &123\ \ \textbf{96}\ 22.0   &117\  \ \textbf{93}\ 20.5\\
$J=12$ &112\ \ \textbf{96}\ 14.3    &115\  \textbf{100}\ 13.0 &107\ \ \textbf{95}\ 11.2  &112\ \ \textbf{95}\ 15.2\\
$J=14$ &110\ \  \textbf{96}\ 12.7  &113\ \  \textbf{98}\ 13.3  &113\ \ \textbf{94}\  16.8  &110\ \  \textbf{89}\ 19.1\\
$J=16$ &111\ \ \textbf{97}\ 12.6   &113\ \  \textbf{94}\ 16.8  &109\ \ \textbf{93}\  14.7  &109\ \ \textbf{95}\ 12.8\\
$J=18$ &114\ \ \textbf{95}\ 16.7  &112\ \ \textbf{92}\  17.9  &111\  \ \textbf{96}\ 13.5  &117\ \ \textbf{93}\ 20.5\\
$J=20$ &113\ \ \textbf{95}\ 15.9  &113\ \ \textbf{97}\  14.2  &115\ \ \textbf{93}\ 19.1  &113\  \ \textbf{89}\ 21.2
\\\hline\hline
\end{tabular}}}
\end{center}
\label{tab_Letter}
\end{table}

%\clearpage

\vspace{0.5in}

\subsection{Experiments on the {\em  Pendigits} Dataset}

\begin{table}[h!]
\caption{\textbf{\em Pendigits}. The test mis-classification errors of  {\em logitboost} and \textbf{\em abc-logitboost}, along with the relative improvement $R_{err}$ ($\%$). }
\begin{center}
{\small
{\begin{tabular}{l l l l l }
\hline \hline
  &$\nu = 0.04$ &$\nu=0.06$ &$\nu=0.08$ &$\nu=0.1$ \\
\hline
$J=4$ &119\ \textbf{92}\ 22.7    &120\  \textbf{93}\  22.5 &118\ \textbf{90}\ 23.7   &119\  \textbf{92}\ 22.7\\
$J=6$ &111\ \textbf{98}\ 11.7    &111\  \textbf{97}\ 12.6 &111\ \textbf{96}\ 13.5   &107\  \textbf{93}\ 13.1\\
$J=8$ &116\ \textbf{97}\ 16.4   &117\  \textbf{94}\ 19.7  &115\  \textbf{95}\ 17.4   &114\   \textbf{93}\ 18.4\\
$J=10$ &116\ \textbf{100}\ 13.8  &115\  \textbf{98}\ 14.8 &116\  \textbf{97}\ 16.4   &111\   \textbf{97}\ 12.6\\
$J=12$ &117\  \textbf{98}\ 16.2    &113\  \textbf{98}\ 13.2 \  &113\  \textbf{98}\ 13.3  &114\  \textbf{98}\ 14.0\\
$J=14$ &113\   \textbf{100}\ 11.5  &115\   \textbf{101}\ 12.2  &112\  \textbf{99}\  11.6 &114\   \textbf{98}\ 14.0\\
$J=16$ &112\  \textbf{100}\ 10.7   &118\   \textbf{97}\ 18.8  &112\  \textbf{98}\  12.5  &113\  \textbf{96}\ 15.0\\
$J=18$ &114\  \textbf{102}\ 10.5  &112\  \textbf{97}\  13.4  &109\   \textbf{99}\ \ 9.2  &112\  \textbf{97}\ 13.4\\
$J=20$ &112\  \textbf{106}\ \ 5.4  &116\  \textbf{102}\  12.1  &113\  \textbf{100}\ 11.5  &107\  \textbf{100}\ \ 6.5
\\\hline\hline
\end{tabular}}

}
\end{center}
\label{tab_Pendigits}
\end{table}

\clearpage

\subsection{Experiments on the {\em  Zipcode} Dataset}

\begin{table}[h!]
\caption{\textbf{\em Zipcode}. The test mis-classification errors of  {\em logitboost} and \textbf{\em abc-logitboost}, along with the relative improvement $R_{err}$ ($\%$). }%\vspace{-0.2in}
\begin{center}
{\small
{\begin{tabular}{l l l l l }
\hline \hline
  &$\nu = 0.04$ &$\nu=0.06$ &$\nu=0.08$ &$\nu=0.1$ \\
\hline
$J=4$ &114\ \textbf{111}\ 2.6    &117\  \textbf{108}\ 7.6 &111\ \textbf{114}\ -2.7   &115\  \textbf{107}\ 7.0\\
$J=6$ &109\ \textbf{101}\ 7.3    &107\  \textbf{102}\ 4.6 &106\ \textbf{98}\ 7.5   &110\  \textbf{99}\ 10.0\\
$J=8$ &110\ \textbf{99}\ 10.0   &108\  \textbf{95}\ 12.0  &108\  \textbf{96}\ 11.1   &108\   \textbf{98}\ 9.3\\
$J=10$ &111\ \textbf{97}\ 12.6  &110\  \textbf{94}\ 14.5 &106\  \textbf{97}\ 8.5   &103\   \textbf{94}\ 8.7\\
$J=12$ &111\  \textbf{98}\ 11.7    &112\  \textbf{98}\ 12.5 \  &111\  \textbf{99}\ 10.8  &108\  \textbf{93}\ 13.9\\
$J=14$ &112\   \textbf{100}\ 10.7  &108\   \textbf{99}\ 8.3  &110\  \textbf{97}\  11.8 &114\   \textbf{92}\ 19.3\\
$J=16$ &111\  \textbf{98}\ 11.7   &114\   \textbf{95}\ 16.7  &110\  \textbf{99}\  10.0  &111\  \textbf{98}\ 11.7\\
$J=18$ &112\  \textbf{96}\ 14.2  &114\  \textbf{98}\  14.0  &109\   \textbf{101}\ \ 7.3  &113\  \textbf{98}\ 13.3\\
$J=20$ &114\  \textbf{97}\ \ 14.9  &108\  \textbf{96}\  11.1  &109\  \textbf{100}\ 8.3  &116\  \textbf{96}\ \ 17.2
\\\hline\hline
\end{tabular}}
}
\end{center}
\label{tab_Zipcode}
\end{table}

\vspace{0.5in}

\subsection{Experiments on the {\em  Optdigits} Dataset}

\begin{table}[h!]
\caption{\textbf{\em Optdigits}.The test mis-classification errors of  {\em logitboost} and \textbf{\em abc-logitboost}, along with the relative improvement $R_{err}$ ($\%$). }
\begin{center}
{\small
{\begin{tabular}{l l l l l }
\hline\hline
  &$\nu = 0.04$ &$\nu=0.06$ &$\nu=0.08$ &$\nu=0.1$ \\\hline
$J=4$ &52\ \textbf{41}\ 21.2    &50\  \textbf{42}\ 16.0 &50\ \textbf{40}\  20.0   &49\  \textbf{41}\ 16.3\\
$J=6$ &52\ \textbf{43}\ 17.3    &52\  \textbf{45}\ 13.5 &53\ \textbf{44}\ 17.0   &52\  \textbf{38}\ 26.9\\
$J=8$ &55\ \textbf{44}\ 20.0   &55\  \textbf{44}\ 20.0 &53\  \textbf{45}\ 15.1   &54\   \textbf{45}\ 16.7\\
$J=10$ &57\ \textbf{50}\ 12.3  &56\  \textbf{50}\ 10.7 &54\  \textbf{46}\  14.8  &55\   \textbf{42}\ 23.6\\
$J=12$ &52\  \textbf{50}\ 3.8    &55\  \textbf{48}\ 12.7 \  &54\  \textbf{47}\ 13.0  &54\  \textbf{46}\ 14.8\\
$J=14$ &58\   \textbf{48}\ 17.2  &55\   \textbf{46}\ 16.4  &56\  \textbf{51}\  8.9 &53\   \textbf{48}\ 9.4\\
$J=16$ &61\  \textbf{54}\ 11.5   &57\   \textbf{51}\ 10.5  &58\  \textbf{49}\  15.5  &56\  \textbf{46}\ 17.9\\
$J=18$ &65\  \textbf{54}\ 16.9  &64\  \textbf{55}\  14.0  &60\   \textbf{53}\  11.7  &66\  \textbf{51}\ 22.7\\
$J=20$ &63\  \textbf{61}\ \ 3.2  &61\  \textbf{56}\  8.2  &64\  \textbf{55}\ 14.1  &64\  \textbf{55}\ \ 14.1
\\\hline\hline
\end{tabular}}
}
\end{center}
\label{tab_Optdigits}
\end{table}

\clearpage

\subsection{Experiments on the {\em  Isolet} Dataset}

For this dataset, \cite{Proc:ABC_ICML09} only experimented with $\nu = 0.1$ for {\em mart} and {\em abc-mart}. We add the experiment results for $\nu = 0.06$.

\begin{table}[h!]
\caption{\textbf{\em Isolet}. The test mis-classification errors of  {\em logitboost} and \textbf{\em abc-logitboost}, along with the relative improvement $R_{err}$ ($\%$). }%\vspace{-0.2in}
\begin{center}
{\small
{\begin{tabular}{l l l}
\hline \hline
  &$\nu = 0.06$ &$\nu=0.1$ \\\hline
$J=4$ &65\ \textbf{55}\ 15.4    &62\  \textbf{55}\ 11.3\\
$J=6$ &67\ \textbf{59}\ 11.9    &69\  \textbf{58}\ 15.9\\
$J=8$ &72\ \textbf{57}\ 20.8   &72\  \textbf{60}\ 16.7 \\
$J=10$ &73\ \textbf{61}\ 16.4  &75\  \textbf{62}\ 17.3 \\
$J=12$ &75\  \textbf{63}\ 16.0    &75\  \textbf{64}\ 14.7 \\
$J=14$ &74\   \textbf{65}\ 12.2  &75\   \textbf{60}\ 20.0  \\
$J=16$ &70\  \textbf{64}\ \ 8.6   &71\   \textbf{62}\ 12.7  \\
$J=18$ &74\  \textbf{67}\ \ 9.5  &73\  \textbf{62}\  15.1  \\
$J=20$ &71\  \textbf{63}\ \ 11.3  &73\  \textbf{65}\  11.0
\\\hline\hline
\end{tabular}}}
\end{center}
\label{tab_Isolet}
\end{table}

\begin{table}[h!]
\caption{\textbf{\em Isolet}. The test mis-classification errors of  {\em mart} and \textbf{\em abc-mart}, along with the relative improvement $R_{err}$ ($\%$). }
\begin{center}
{\small
{\begin{tabular}{l l l}
\hline \hline
  &$\nu = 0.06$ &$\nu=0.1$ \\\hline
$J=4$ &81\ \textbf{68}\ 16.1    &80\  \textbf{64}\ 20.0\\
$J=6$ &86\ \textbf{71}\ 17.4    &84\  \textbf{67}\ 20.2\\
$J=8$ &86\ \textbf{72}\ 16.3   &84\  \textbf{72}\ 14.3\\
$J=10$ &87\ \textbf{74}\ 14.9  &82\  \textbf{74}\ \ 9.8\\
$J=12$ &93\  \textbf{73}\ 21.5    &91\  \textbf{74}\ 18.7 \\
$J=14$ &92\   \textbf{73}\ 20.7  &95\   \textbf{74}\ 22.1 \\
$J=16$ &91\  \textbf{73}\ \ 19.8   &94\   \textbf{78}\ 17.0  \\
$J=18$ &86\  \textbf{75}\ \ 12.8  &86\  \textbf{78}\  \ 9.3 \\
$J=20$ &95\  \textbf{79}\ \ 16.8  &87\  \textbf{78}\  10.3
\\\hline\hline
\end{tabular}}}
\end{center}
\label{tab_Isolet_mart}
\end{table}

\vspace{0.2in}

\section{Conclusion}

Multi-class classification is a fundamental task in machine learning. This paper presents the {\em abc-logitboost} algorithm and demonstrates its considerable improvements over {\em logitboost} and {\em abc-mart} on a variety of datasets.\\

There is one interesting UCI dataset named {\em Poker}, with 25K training samples and 1 million testing samples. Our experiments showed that {\em abc-boost} could achieve an accuracy $>90\%$ (i.e., the error rate $<10\%$). Interestingly, using LibSVM, an accuracy of about $60\%$ was obtained\footnote{Chih-Jen Lin. Private communications in May 2009 and August 2009.}. We will report the results in a separate paper.

%\bibliographystyle{plain}
%\bibliography{../bib/IEEEabrv,../bib/mybibfile}

\begin{thebibliography}{10}

\bibitem{Book:Agresti}
Alan Agresti.
\newblock {\em Categorical Data Analysis}.
\newblock John Wiley \& Sons, Inc., Hoboken, NJ, second edition, 2002.

\bibitem{Article:Bartlett_AS98}
Peter Bartlett, Yoav Freund, Wee~Sun Lee, and Robert~E. Schapire.
\newblock Boosting the margin: a new explanation for the effectiveness of
  voting methods.
\newblock {\em The Annals of Statistics}, 26(5):1651--1686, 1998.

\bibitem{Article:Begg_84}
Colin~B. Begg and Robert Gray.
\newblock Calculation of polychotomous logistic regression parameters using
  individualized regressions.
\newblock {\em Biometrika}, 71(1):11--18, 1984.

\bibitem{Article:Freund_95}
Yoav Freund.
\newblock Boosting a weak learning algorithm by majority.
\newblock {\em Inf. Comput.}, 121(2):256--285, 1995.

\bibitem{Article:Freund_JCSS97}
Yoav Freund and Robert~E. Schapire.
\newblock A decision-theoretic generalization of on-line learning and an
  application to boosting.
\newblock {\em J. Comput. Syst. Sci.}, 55(1):119--139, 1997.

\bibitem{Article:Friedman_AS01}
Jerome~H. Friedman.
\newblock Greedy function approximation: A gradient boosting machine.
\newblock {\em The Annals of Statistics}, 29(5):1189--1232, 2001.

\bibitem{Article:FHT_AS00}
Jerome~H. Friedman, Trevor~J. Hastie, and Robert Tibshirani.
\newblock Additive logistic regression: a statistical view of boosting.
\newblock {\em The Annals of Statistics}, 28(2):337--407, 2000.

\bibitem{Article:FHT_JMLR08}
Jerome~H. Friedman, Trevor~J. Hastie, and Robert Tibshirani.
\newblock Response to evidence contrary to the statistical view of boosting.
\newblock {\em Journal of Machine Learning Research}, 9:175--180, 2008.

\bibitem{Article:Lee_JASA04}
Yoonkyung Lee, Yi~Lin, and Grace Wahba.
\newblock Multicategory support vector machines: Theory and application to the
  classification of microarray data and satellite radiance data.
\newblock {\em Journal of the American Statistical Association},
  99(465):67--81, 2004.

\bibitem{Proc:ABC_ICML09}
Ping Li.
\newblock Abc-boost: Adaptive base class boost for multi-class classification.
\newblock In {\em ICML}, Montreal, Canada, 2009.

\bibitem{Report:Li_Robust-LogitBoost}
Ping Li.
\newblock Robust logitboost.
\newblock Technical report, Department of Statistical Science, Cornell
  University, 2009.

\bibitem{Proc:McRank_NIPS07}
Ping Li, Christopher~J.C. Burges, and Qiang Wu.
\newblock Mcrank: Learning to rank using classification and gradient boosting.
\newblock In {\em NIPS}, Vancouver, BC, Canada, 2008.

\bibitem{Proc:Mason_NIPS00}
Liew Mason, Jonathan Baxter, Peter Bartlett, and Marcus Frean.
\newblock Boosting algorithms as gradient descent.
\newblock In {\em NIPS}, 2000.

\bibitem{Article:Schapire_ML90}
Robert Schapire.
\newblock The strength of weak learnability.
\newblock {\em Machine Learning}, 5(2):197--227, 1990.

\bibitem{Article:Schapire_ML99}
Robert~E. Schapire and Yoram Singer.
\newblock Improved boosting algorithms using confidence-rated predictions.
\newblock {\em Machine Learning}, 37(3):297--336, 1999.

\bibitem{Article:Tewari_JMLR07}
Ambuj Tewari and Peter~L. Bartlett.
\newblock On the consistency of multiclass classification methods.
\newblock {\em Journal of Machine Learning Research}, 8:1007--1025, 2007.

\bibitem{Article:Zhang_JMLR04}
Tong Zhang.
\newblock Statistical analysis of some multi-category large margin
  classification methods.
\newblock {\em Journal of Machine Learning Research}, 5:1225--1251, 2004.

\bibitem{Article:Zou_AOAS08}
Hui Zou, Ji~Zhu, and Trevor Hastie.
\newblock New multicategory boosting algorithms based on multicategory
  fisher-consistent losses.
\newblock {\em The Annals of Applied Statistics}, 2(4):1290--1306, 2008.

\end{thebibliography}

\end{document}